\documentclass[times, review, 10pt]{elsarticle}

\usepackage{amssymb}
\usepackage{amsmath}

\usepackage{graphicx} 
\usepackage{amsfonts}
\usepackage{amssymb}
\usepackage{amsthm}
\usepackage{amsmath}
\usepackage{mathrsfs}
\usepackage{xcolor}
\usepackage{booktabs}
\usepackage{algorithm}
\usepackage{algpseudocode}
\usepackage{booktabs}
\usepackage{array}
\usepackage{caption}
\usepackage{subcaption}
\usepackage{longtable}
\usepackage{adjustbox}
\usepackage{bm}  
\usepackage{hyperref}
\usepackage{lscape}

\usepackage{siunitx}
\newcommand{\R}{\mathbb{R}}

\newcommand{\lb}{\left (}
\newcommand{\rb}{\right )}

\newcommand\eqn[1]{\begin{align*}#1\end{align*}}
\newcommand\eqnn[1]{\begin{align}#1\end{align}}

\newcommand{\argmin}{\textrm{argmin}}

\newtheorem{proposition}{Proposition}

\newtheorem{theorem}{Theorem}

\journal{Pattern Recognition}

\begin{document}

\begin{frontmatter}

\title{Manifold Dimension Estimation via Local Graph Structure} 

\author[unsw]{Bi, Zelong\corref{cor1}}\ead{zelong.bi@unsw.edu.au}
\author[unsw]{{Lafaye De Micheaux}, Pierre}\ead{lafaye@unsw.edu.au}
\cortext[cor1]{Corresponding author.}

\affiliation[unsw]{
  organization={School of Mathematics and Statistics, University of New South Wales},
  addressline={High St},
  city={Kensington},
  postcode={2052},
  state={NSW},
  country={Australia}
}

\begin{abstract}

Most existing manifold dimension estimators rely on the assumption that the underlying manifold is locally flat within the neighborhoods under consideration. More recently, curvature-adjusted principal component analysis (\textbf{CA-PCA}) has emerged as a powerful alternative by explicitly accounting for the manifold’s curvature. Motivated by these ideas, we propose a manifold dimension estimation framework that captures the local graph structure of the manifold through regression on local PCA coordinates. Within this framework, we introduce two representative estimators: quadratic embedding (\textbf{QE}) and total least squares (\textbf{TLS}). Experiments on both synthetic and real-world datasets demonstrate that these methods perform competitively with, and often outperform, state-of-the-art approaches.
\end{abstract}


\end{frontmatter}

\section{Introduction}

Representing high-dimensional data using a substantially smaller number of variables is an important task in many modern machine learning procedures. Low-dimensional representations improve computational efficiency, enhance interpretability, and provide a potential explanation for why learning algorithms often succeed despite the curse of dimensionality \citep{jia2022feature}.

Classical dimensionality reduction methods, such as principal component analysis (PCA; \citep{wold1987principal}), assume that the data lie in a linear subspace of the ambient space $\mathbb{R}^p$. Under this assumption, a single global linear projection yields a faithful low-dimensional representation. However, this assumption might not hold for many datasets in practice.

A more flexible perspective is provided by the \emph{manifold hypothesis}, which posits that high-dimensional data points lie on or close to a smooth $d$-dimensional manifold embedded in $\mathbb{R}^p$. Although a global coordinate system may not exist, the data can still be locally described using only $d$ coordinates. Recovering this underlying manifold structure, and thereby obtaining local low-dimensional representations, is referred to as \emph{manifold learning} \citep{meilua2024manifold}. Classical manifold learning methods include Isomap \citep{tenenbaum2000global}, diffusion map \citep{coifman2006diffusion}, and UMAP \citep{mcinnes2018umap}, typically designed to preserve certain geometric properties between high- and low-dimensional representations. More recent techniques, such as those in \citep{alberti2024manifold}, often employ variational autoencoder (VAE; \citep{kingma2013auto}) to obtain representations that more faithfully respect the underlying geometry. 

Leveraging low-dimensional representations has led to notable improvements in many downstream tasks. For example, in \citep{li2026hy}, a hybrid facial expression classification framework incorporating UMAP for feature reduction achieves higher accuracy than baseline and conventional configurations. Moreover, in generative settings, recent work suggests that diffusion models \citep{ho2020denoising} provide an implicit representation of the data manifold \citep{park2023understanding}. And constructing diffusion models in latent spaces has become an active research direction \citep{rombach2022high, liang2025latent}, which requires a solid understanding of the underlying manifold.

Almost all manifold learning techniques assume that the intrinsic dimension $d$ of the underlying manifold is known a priori, which is often not the case in practical settings. Consequently, a closely related problem in nonlinear dimensionality reduction is \emph{manifold dimension estimation}, which aims to estimate $d$ directly from data. Estimating the intrinsic dimension of the underlying manifold is a challenging problem in its own right, and a variety of methods have been proposed over the years.

The majority of manifold dimension estimators—for example, the local PCA estimator (\textbf{Local PCA}, \citep{fukunaga1971algorithm, fan2010intrinsic}), the maximum likelihood estimator (\textbf{MLE}, \citep{levina2004maximum}), the dimensionality estimator based on angle and norm concentration (\textbf{DanCo}, \citep{CERUTI20142569}), the manifold-adaptive dimension estimator (\textbf{MADA}, \citep{farahmand2007manifold}), the intrinsic dimension estimator with tight localities (\textbf{TLE}, \citep{amsaleg2019intrinsic}), and the two-nearest neighbor estimator (\textbf{TwoNN}, \citep{facco2017estimating})—rely on the so-called \emph{flatness assumption}, whereby sufficiently small neighborhoods of the manifold are approximated by $d$-dimensional linear subspaces. This simplification transforms the nonlinear geometric estimation problem into a linear one, enabling the design for a wide range of estimators.

Other representative estimators include the one based on the Wasserstein distance (\textbf{Wasserstein}, \citep{block2022intrinsic}), which exploits the fact that the rate at which the empirical distribution converges to the true distribution in Wasserstein distance scales with the intrinsic dimension $d$ rather than the ambient dimension $p$ \citep{dudley1969speed}. More recently, the curvature-adjusted PCA estimator (\textbf{CA-PCA}, \citep{gilbert2025pca}) has been proposed, which explicitly incorporates non-flat geometric information into the local PCA procedure by correcting the eigenstructure via curvature adjustments. It has been shown to substantially improve upon the already powerful but often downplayed \textbf{Local PCA} estimator\footnote{\textcolor{black}{A brief introduction to \textbf{CA-PCA} procedure can be found in Appendix E}}.

Despite substantial effort, previous surveys and empirical studies \citep{campadelli2015intrinsic, camastra2016intrinsic}, including our most recent work \citep{bi2025manifold}, have shown that the problem of manifold dimension estimation remains far from being fully resolved, with considerable room for improvement. Among the estimators introduced above, \textbf{DanCo}, \textbf{TwoNN}, and \textbf{CA-PCA} are generally identified as overall top performers. However, \textbf{DanCo} is computationally expensive; \textbf{TwoNN} may underestimate the intrinsic dimension in high-dimensional settings; and \textbf{CA-PCA} tends to largely overestimate the dimension when the manifold is nonlinearly embedded in an ambient space with much higher dimensions, a limitation also shared by \textbf{Local PCA} and arising from their eigenvalue-based comparison procedure.


Motivated by both the strengths and limitations of existing estimators—particularly the robustness of PCA-based approaches and the improvements achieved by \textbf{CA-PCA} through its explicit incorporation of curvature—we propose a new regression-based framework for manifold dimension estimation. In contrast to \textbf{CA-PCA}, our method does not rely on explicit curvature estimation or eigenvalue correction. Instead, we recover the local graph structure of the manifold by regressing normal components onto tangent components obtained via PCA within each neighborhood.


The regression-based formulation has two main advantages. First, noise and deviations can be naturally incorporated into the error term of the model specification, allowing our estimators to capture structure from imperfect, noisy data. In comparison, many existing estimators, including \textbf{CA-PCA}, are designed under the assumption of noiseless data, making them less suited for real-world data where the manifold hypothesis is unlikely to hold exactly. {\color{black}Second, regression provides a natural framework for decision making and theoretical analysis. In particular, we establish theoretical properties of the proposed estimators which provide probabilistic guarantees for their performance. Extensive experiments further demonstrate their competitive or improved estimation accuracy over state-of-the-art methods, while effectively mitigating the overestimation issue of \textbf{CA-PCA} and retaining the benefits of incorporating curvature information.}

Our main contributions are as follows:

\begin{itemize}
    \item \textbf{Regression-based formulation.} We introduce a new framework that casts manifold dimension estimation as a recursive regression problem, \textcolor{black}{enabling the modeling of both complex geometry as well as noise and deviations}. To the best of our knowledge, this perspective is novel in the manifold dimension estimation literature.

    \item \textbf{Representative estimators.} We instantiate the framework with two simple yet effective methods: quadratic embedding (\textbf{QE}), a second-order regression model that captures curvature effects, and total least squares (\textbf{TLS}), which further accounts for errors in both predictors and responses.

    \item {\color{black}\textbf{Theoretical and empirical validation.} We establish theoretical properties of our estimators and demonstrate their robustness and accuracy through extensive experiments on synthetic and real-world datasets, particularly in small-sample regimes and for manifolds with complex curvature. The proposed \textbf{QE} and \textbf{TLS} estimators achieve competitive or superior performance compared to state-of-the-art methods.}
\end{itemize}

The remainder of this article is organized as follows. Section~2 reviews fundamental manifold constructions and introduces our design framework. Section~3 presents the two proposed estimators in details, followed by experimental results and their discussion in Section~4. We conclude in Section~5.

\section{Manifold Geometry}

In this section, we briefly review several fundamental manifold constructions essential for understanding the methods introduced in this article, before presenting our general design framework for manifold dimension estimation. A more comprehensive treatment of the theoretical foundation is provided in~\citep{bi2025manifold}. For readers interested in more advanced topics in differential geometry, we refer to \citep{lee2003smooth} and \citep{lee2018introduction}, which offer detailed expositions of the relevant mathematical background. To facilitate readability, a summary of notation is provided in Table~\ref{tab:notation}.

\subsection{Tangent Space and Local Graph Representation}

\begin{figure}[H]
    \centering
    \begin{subfigure}[b]{0.45\textwidth}
        \centering
        \includegraphics[width=\textwidth]{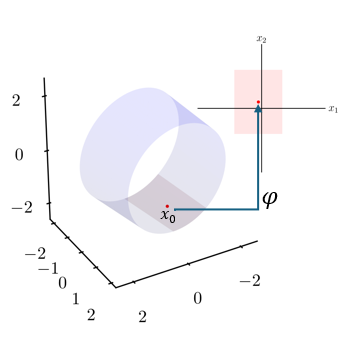}
        \caption{Local coordinates around $\bm{x}_0$.}
        \label{fig:chart}
    \end{subfigure}
    \hfill
    \begin{subfigure}[b]{0.45\textwidth}
        \centering
        \includegraphics[width=\textwidth]{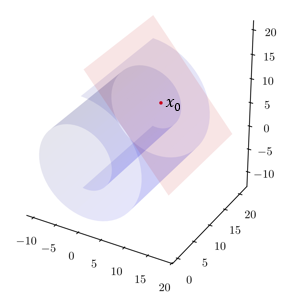}
        \caption{Tangent space at $\bm{x}_0$.}
        \label{fig:tangent}
    \end{subfigure}
    \caption{Local coordinate and tangent space.}
\end{figure}

Throughout this article, we take a manifold \( M \) as a \( d \)-dimensional Riemannian submanifold embedded in ambient space \( \mathbb{R}^p \), generalizing smooth curves and surfaces to higher dimensions. The intrinsic dimension \( d \) then refers to the number of free parameters required to describe the manifold locally. That is, for any point \( \bm{x}_0 \in M \), there exists a neighborhood \( U(\bm{x}_0) \subseteq M \) that can be smoothly mapped to \( \mathbb{R}^d \) via an injective function \( \varphi\), as illustrated in Figure~\ref{fig:chart}.

Let \( \varphi^{-1} \) denote the inverse of \( \varphi \). The \emph{tangent space} of \( M \) at \( \bm{x}_0 \) is defined as the \( d \)-dimensional vector space (with operations trivially defined with respect to $\bm{x}_0$) given by
\[
T_{\bm{x}_0}M = \{ \bm{x}_0 + d\varphi^{-1}\big|_{\varphi(\bm{x}_0)}(\bm{u}) : \bm{u} \in \mathbb{R}^d \}.
\]
Here, \( \bm{x}_0 + d\varphi^{-1}\big|_{\varphi(\bm{x}_0)}(\bm{u}) \) corresponds to the first-order Taylor expansion of \( \varphi^{-1} \), making \( T_{\bm{x}_0}M \) the best linear approximation of the manifold near \( \bm{x}_0 \), as illustrated in Figure~\ref{fig:tangent}.

The tangent space approximation forms the basis of the widely used \emph{flatness assumption} in manifold dimension estimation. Suppose we are given a data point \( \bm{x}_k \) along with its \( K \)-nearest neighbors sampled from \( M \). The flatness assumption ignores any geometric or distributional irregularities in the neighborhood and posits the data are uniformly distributed within a \( d \)-dimensional ball \( B_{\bm{x}_k}(R) \subseteq T_{\bm{x}_k}M \). This local linearization enables the construction of various statistics related to \( d \). In this context, “flatness” refers to ignoring curvature at a local scale.

\begin{figure}[H]
    \centering
    \begin{subfigure}[b]{0.43\textwidth}
        \centering
        \includegraphics[width=\textwidth]{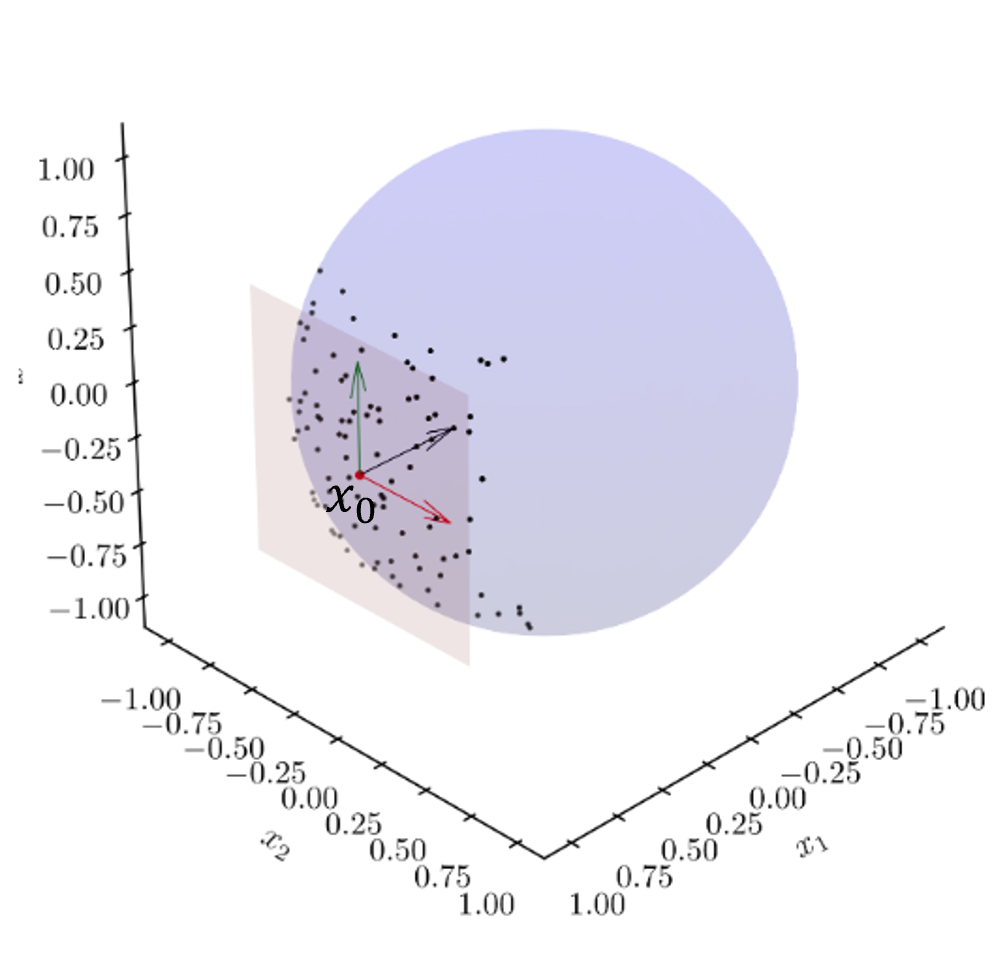}
        \caption{Linear Approximation around $\bm{x}_0$.}
        \label{fig:la}
    \end{subfigure}
    \hfill
    \begin{subfigure}[b]{0.45\textwidth}
        \centering
        \includegraphics[width=\textwidth]{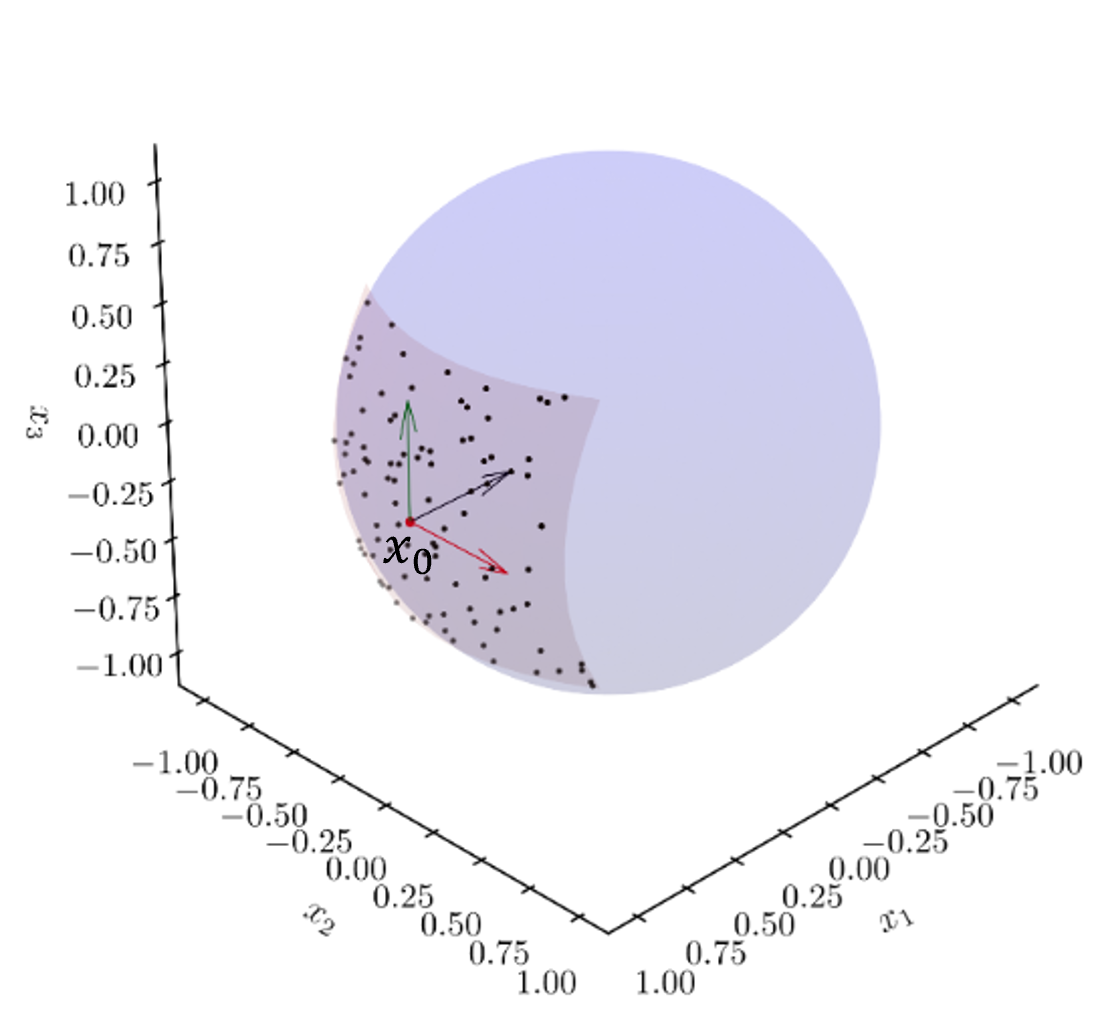}
        \caption{Quadratic Approximation around $\bm{x}_0$.}
        \label{fig:qa}
    \end{subfigure}
    \caption{Local Graph Representation.}
\end{figure}

Furthermore, the neighborhood around \( \bm{x}_0 \) can be reparameterized to better reflect the manifold's local geometry. By translating \( \bm{x}_0 \) to the origin, we can construct a new coordinate system consisting of \( d \) orthonormal vectors spanning the tangent space \( T_{\bm{x}_0}M \), and \( p - d \) orthonormal vectors spanning the normal space \( (T_{\bm{x}_0}M)^\perp \), as shown in Figure~\ref{fig:la}. Standard results from differential geometry \citep{lee2003smooth} then guarantee that any point \( \bm{x} \) sufficiently close to \( \bm{x}_0 \) can be expressed in these new coordinates as
\begin{equation}\label{eqn-lgr}
\bm{x}' = \big(\bm{x}'_{1:d},\, \bm{g}(\bm{x}'_{1:d})\big),
\end{equation}
where \( \bm{x}'_{1:d}=(x'_1,\ldots,x'_d)^\top \), \( \bm{x}'_0 = \bm{0} \), and \( \bm{g} \colon \mathbb{R}^d \to \mathbb{R}^{p-d} \) is a smooth function satisfying \( \nabla \bm{g}(\bm{0}) = \bm{0} \), which we refer to as the \textit{local graph function}.

In this new coordinate system, the neighborhood of \( \bm{x}_0 \) is represented as the graph of \( \bm{g} \). The first \( d \) components, \( \bm{x}'_{1:d} \), serve as coordinates in the tangent space, while the remaining \( p - d \) components are determined by~\( \bm{g} \) and capture deviations from the tangent space approximation. The \emph{flatness assumption} corresponds to the case where \( \bm{g} \doteq 0 \), meaning the last \( p - d \) coordinates vanish—equivalent to retaining only the first-order Taylor expansion of \( \bm{g} \) around the origin. Consequently, the nontrivial local geometry of the manifold is entirely encoded in the higher-order terms of the Taylor expansion of \( \bm{g} \). In particular, the second-order Taylor polynomial of \( \bm{g} \) provides a quadratic approximation, as illustrated in Figure~\ref{fig:qa}.

\subsection{Linear and Nonlinear Embedding}

A particularly challenging scenario for manifold dimension estimators arises when the underlying manifold \( M \) is \emph{space-filling} or \emph{nonlinearly embedded} in the ambient space \citep{campadelli2015intrinsic}. Such manifolds typically exhibit more complex curvature. According to the local graph representation discussed above, since there are \( d \) linearly independent directions in the tangent space and \( p - d \) linearly independent normal directions, a total of \( d(p - d) \) values are required to fully describe the curvature at any given point. The more complex the curvature, the more data are needed to accurately capture the underlying geometry, making dimension estimation increasingly difficult.

For example, a 3-dimensional sphere linearly embedded in \( \mathbb{R}^{6} \) is actually contained within a 4-dimensional subspace, and thus its curvature vanishes in many directions (i.e., $g_2$ and $g_3$ are zero, where the \( g_j \) are scalar component functions of \( \bm{g} \)). In contrast, consider a 3-dimensional \emph{deformed sphere} embedded in \( \mathbb{R}^{6} \), defined via the mapping
\begin{equation}
    \varphi^{-1}(\bm{u}) = (x_1, \ldots, x_{6}),
\end{equation}
with
\begin{equation}
x_j     = \left[R + r \cos(2c\pi u_j)\right] \cos(2\pi u_j), \quad
x_{j+3} = \left[R + r \cos(2c\pi u_j)\right] \sin(2\pi u_j), 
\end{equation}

for \( j = 1, \ldots, 3 \), where \( R, r, c > 0 \) are fixed constants. This manifold exhibits nontrivial curvature in all \( 3 \times 3 \) tangent and normal directions. Estimating its intrinsic dimension is therefore significantly more difficult due to the pervasive and nonlinear curvature present in all local neighborhoods.

A major weakness of flatness-based estimators such as \textbf{Local PCA} and \textbf{DanCo} is their deteriorated performance on nonlinearly embedded manifolds like the deformed sphere. Given the same number of data points in a neighborhood, the flatness assumption is increasingly violated as the curvature becomes more complex.

\subsection{Capturing Local Graph Structure}

Our framework for manifold dimension estimation leverages the local graph structure of the underlying manifold in a direct and natural way. Let $\{\bm{x}_k\}_{k=1}^n \subseteq \R^p$ be a dataset satisfying the manifold hypothesis, supported on a $d$-dimensional manifold $M$. According to \ref{eqn-lgr}, for each point $\bm{x}_k$ there exists a local graph function $\bm{g}:\mathbb{R}^d \to \mathbb{R}^{p-d}$ that characterizes the graph structure of its neighborhood. Consequently, data points sufficiently close to $\bm{x}_k$, when their components are expressed in new coordinates with respect to $\bm{x}_k$, will also satisfy the relation described by $\bm{g}$, at least approximately. If $\bm{g}$ can be identified, then the intrinsic dimension of the manifold is naturally revealed as the splitting point between its input and output dimensions.

For each data point $\bm{x}_k$, our framework aims to identify the local graph structure in its neighborhood, namely $\bm{g}$, using its nearest neighbors. In practice, neither the intrinsic dimension $d$ nor the functional form of $\bm{g}$ is known a priori. To operationalize this idea, we can consider, for each candidate dimension $j$, a model $\hat{\bm{g}}_j:\R^j \to \R^{p - j}$ designed to approximate $\bm{g}$. We then fit $\bm{x}_k$ and its neighboring points to this model and compute a corresponding goodness-of-fit score $\xi_j$ to quantify how well the model explains the observed data. Ideally, $\hat{\bm{g}}_j$ should be the closest to $\bm{g}$ when $j = d$, resulting in the highest score. Consequently, the intrinsic dimension can be estimated as
\begin{equation}\label{eqn-estimator}
\hat{d} = \arg\max_{1 \leq j \leq p} \xi_j.
\end{equation}

Given that $\bm{g}$ is a smooth function, a natural choice for the approximating models $\hat{\bm{g}}_j$ should correspond to its Taylor expansions. Since $\bm{g}(\bm{0}) = \bm{0}$ and $\nabla \bm{g}(\bm{0}) = \bm{0}$ (under the new coordinates), capturing meaningful local geometry requires incorporating at least second-order terms. This leads to the following quadratic form for each component of $\hat{\bm{g}}_j$:
\begin{equation}\label{eqn-qa-tensor}
   q_{j\ell}(\bm{x}_{1:j}') = \bm{x}_{1:j}'^\top Q_{j\ell} \bm{x}_{1:j}', \quad \ell = 1, \ldots, p - j,
\end{equation}
where each $Q_{j\ell}$ is a $j \times j$ symmetric matrix whose entries correspond to the second-order derivatives of $\bm{g}$ (with respect to the $\ell$-th output) and encode local curvature information. This will be the model class employed by our proposed estimators in the next section.

Further improvements are possible when additional geometric information about the underlying manifold is available. For instance, if the manifold is known to be a $d$-dimensional sphere of radius $R$ (with $d$, and possibly $R$, unknown), we can exploit the fact that the local graph function admits a common structure across neighborhoods:
\begin{equation}
   \bm{g}(\bm{x}_{1:d}') = \left( R - \sqrt{R^2 - \|\bm{x}_{1:d}'\|^2},\, 0, \ldots, 0 \right),
\end{equation}
where the first nonzero component captures all curvature of the sphere rather than just being a second-order approximation. Incorporating such prior geometric information can yield more accurate models which can potentially improve dimension estimation.

To summarize, our proposed framework for manifold dimension estimation focuses on accurately capturing the local graph structure of the underlying manifold. For each candidate dimension, specific local graph models are specified, possibly guided by prior geometric knowledge. These models are then fitted to the data, and their goodness of fit is evaluated, the intrinsic dimension is estimated as the dimension corresponding to the best fit.

Our framework offers several advantages over existing approaches:
\begin{itemize}
    \item \textbf{Curvature-aware modeling.} By allowing flexible specification of local graph functions—particularly through quadratic forms—we explicitly model the curvature of the underlying manifold. Unlike methods such as \textbf{CA-PCA}, which attempt to infer curvature directly from sparse and potentially noisy local samples, our approach embeds the problem into a regression framework, transforming it into a more stable model-fitting task.

    \item \textbf{Flexibility and robustness.} The framework allows flexible model choices across candidate dimensions and neighborhoods, enabling the incorporation of prior geometric knowledge. While quadratic models serve as a natural default, more expressive alternatives (e.g., semiparametric or neural network models \citep{devore2021neural}) can be used, though they may be impractical given limited local sample sizes. 

    \item \textbf{Robustness to noise.} \textcolor{black}{On top of the manifold structure encoded in $\bm{g}$, our framework can naturally accommodate noise and deviations through the error term in regression models; i.e., in a given neighborhood, the data can be described as
    \eqn{
        \bm{x}_{(d+1):p}' = \bm{g}(\bm{x}_{1:d}') + \bm{\varepsilon}.
    }
    The error term $\bm{\varepsilon}$ incorporates both deviations from the function form we specify for $\bm{g}$, as well as other types of perturbations that push data points away from the underlying manifold.
    }
    
\end{itemize}

\section{Design of New Estimators}\label{sec-3}

The concepts behind \textbf{QE} and \textbf{TLS} follow naturally from our framework. For both estimators, the model specification for candidate dimensions up to \( p - 1 \) is given by the quadratic form in \eqref{eqn-qa-tensor}. With new coordinates for each neighborhood in hand, the model for dimension \( j \) can be fitted by treating the first \( j \) coordinates of each data point as inputs and the remaining \( p - j \) coordinates as outputs. A local PCA procedure is applied to each neighborhood to estimate the new coordinates guaranteed by the theory, which can be reused across all candidate dimensions. The two estimators differ in their choice of fitting and evaluation methods: \textbf{QE} uses ordinary least squares (OLS), while \textbf{TLS} employs total least squares (TLS, \citep{golub1980analysis}), which accounts for deviation and noise in both input and output directions.

\subsection{Regression with PCA Coordinates}

Following the above discussion, given a data point \(\bm{x}_k\in\{\bm{x}_1,\ldots,\bm{x}_n\}\) 
and its \(K\)-nearest neighbors \(\bm{x}_k^1, \ldots, \bm{x}_k^K\) (ranked by their Euclidean distances from \(\bm{x}_k\)), we first apply PCA to compute the eigendecomposition of the sample covariance matrix of these $K+1$ points (after centering them at their local mean, $\bar{\bm{x}}_k)$. This yields an orthonormal basis whose vectors \(\bm{v}_1, \ldots, \bm{v}_p\) are stored in matrix $V$, with corresponding eigenvalues \(\lambda_1 \geq \cdots \geq \lambda_p\). Taking the local mean \(\bar{\bm{x}}_k\) as the new origin, each point in the neighborhood can then be expressed in a new coordinate system aligned with the principal directions, denoted by \(\bm{x}_0', \ldots, \bm{x}_K'\), which are given by
\eqnn{
\bm{x}_0' = V^\top(\bm{x}_k - \bar{\bm{x}}_k),\quad\bm{x}_\ell' = V^\top(\bm{x}_k^\ell - \bar{\bm{x}}_k), \ \ \ \ell = 1, \cdots, K.
}

\begin{figure}[H]
    \centering
    \begin{subfigure}[b]{0.45\textwidth}
        \centering
        \includegraphics[width=\textwidth]{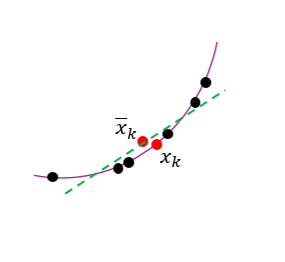}
        \caption{Local linear approximation via PCA.}
        \label{fig:approxpca}
    \end{subfigure}
    \hfill
    \begin{subfigure}[b]{0.45\textwidth}
        \centering
        \includegraphics[width=\textwidth]{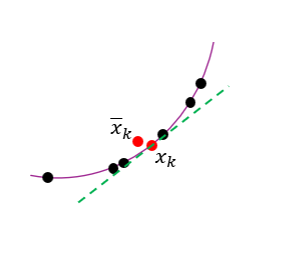}
        \caption{Local linear approximation via \(T_{\bm{x}_k}M\).}
        \label{fig:approxtangent}
    \end{subfigure}
    \caption{Practical and theoretical local approximations.}
    \label{fig:twoapprox}
\end{figure}

By construction, the subspace spanned by \(\bm{v}_1, \ldots, \bm{v}_d\) (centered at \(\bar{\bm{x}}_k\)) provides the optimal \(d\)-dimensional linear approximation of the data points, in the sense that it minimizes the sum of squared distances from the points to the subspace, among all linear spaces of dimension at most \(d\) \citep{greenacre2022principal}. This yields a practical linear approximation of the data points, as shown in Figure~\ref{fig:approxpca}.

On the theoretical side, {\color{black} assuming $\bm{x}_k$ lies on $M$,} it is also known that the entire neighborhood around \(\bm{x}_k\) can be approximated by the tangent space \(T_{\bm{x}_k}M\). Under the local graph representation (Eq.~\eqref{eqn-lgr}), this corresponds to taking the first-order Taylor approximation of the function \(\bm{g}\). This approximation is illustrated in Figure~\ref{fig:approxtangent}.

{\color{black} Intuitively, the two approximations should be close at sufficiently small scales. Proposition~\ref{prop:population-pca} in Section~\ref{sec-analysis} formally confirms this intuition by showing that, under an approximate local model, the first \(d\) principal directions of population PCA span the tangent space \(T_{\bm{x}_k}M\). Indeed, PCA is widely regarded as the most natural method for tangent space estimation; see \citep{lim2024tangent} for an alternative justification.}

For our purposes, the most important consequence is that PCA coordinates provide accurate empirical estimates of the theoretical coordinates involved in the local graph representation, thereby allowing us to fit the models corresponding to different candidate dimensions.
At its core, our estimator operates by identifying the splitting index $d$ at which a deterministic relationship emerges between the first $d$ coordinates and the remaining $p-d$, governed by the local graph function.
The correct intrinsic dimension corresponds to the point where a marked improvement in model fit occurs, thereby revealing the true value of 
$d$.

Using the PCA coordinates, recall the quadratic model for dimension $j$
\begin{equation}\label{eqn-qa}
   q_{j\ell}(\bm{x}_{1:j}') = \bm{x}_{1:j}'^\top Q_{j\ell} \bm{x}_{1:j}', \quad \ell = 1, \ldots, p - j,
\end{equation}
which can be rewritten as linear models by expanding all quadratic and cross-product terms of the inputs $\bm{x}'_1,\ldots,\bm{x}'_j$. For each neighborhood, multivariate linear models such as 
\begin{equation}
\bm{x}_{(j+1):p}' = (q_{j1}(\bm{x}_{1:j}'),\ldots,q_{j, p-j}(\bm{x}_{1:j}'))^\top + \bm{\epsilon}    
\end{equation}
($j=1,\ldots,p-1$) could then be fitted by treating the last $p-j$ PCA coordinates as response variables and applying ordinary least squares (OLS).

Our \textbf{QE} estimator simplifies this approach by restricting ourselves to univariate linear models 
\begin{equation}\label{eqn-model}
x_{j+1}' = q_{j1}(\bm{x}_{1:j}') + \epsilon,  
\end{equation}
which only takes the first output component of the model. {\color{black}Doing this has several advantages. First, we avoid PCA coordinates associated with smaller eigenvalues, which are typically more sensitive to noise. Second, Proposition~\ref{prop:first-normal-component} in Section~\ref{sec-analysis} shows that, as long as the manifold has nonzero curvature, the first normal component always contains a nonconstant quadratic function of the first \(d\) tangent coordinates, providing a theoretical basis for using \(x'_{d+1}\) as the response. This also helps avoid spurious relationships when \(j<d\), while using all \(p-j\) output coordinates can lead to unnecessary parameters and a combinatorial explosion, especially when \(p\) is large.}

Our TLS-based estimator (\textbf{TLS}) mirrors the structure of \textbf{QE}, with the key distinction that the regression step is performed using TLS instead of OLS. We also experimented with LASSO \citep{tibshirani1996regression} and ridge regression \citep{mcdonald2009ridge} as alternative fitting methods, but found that they yielded inferior performance in this context, which we find unsurprising as the extra penalty terms introduced for these two procedures have no connection with the underlying geometry.

\subsection{Filling Details: QE and TLS}

Evaluating the model fit is a crucial step within our framework. As \eqref{eqn-estimator} suggests, a good fit signals that the intrinsic dimension has likely been reached. For \textbf{QE}, this step is straightforward: classical statistical theory provides the \( F \)-statistic \citep{montgomery2021introduction}, specifically designed to compare a target regression model against others.

\begin{figure}[H]
    \centering

    \begin{subfigure}[t]{0.32\textwidth}
        \centering
        \includegraphics[width=\textwidth]{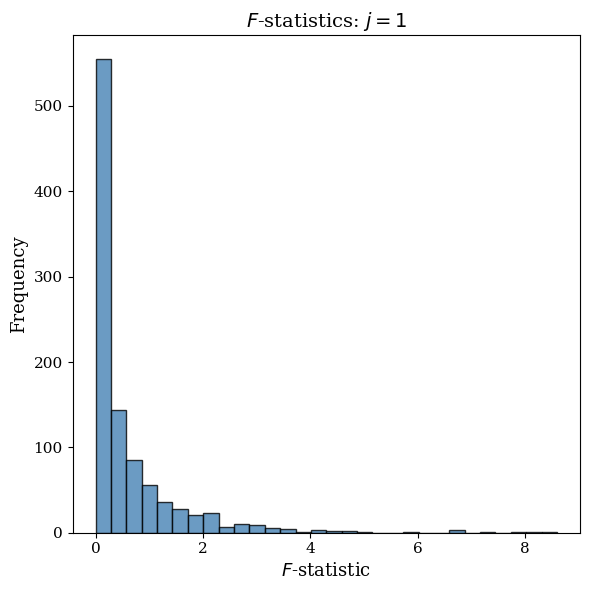}
        \caption{$j=1$}
    \end{subfigure}
    \hfill
    \begin{subfigure}[t]{0.32\textwidth}
        \centering
        \includegraphics[width=\textwidth]{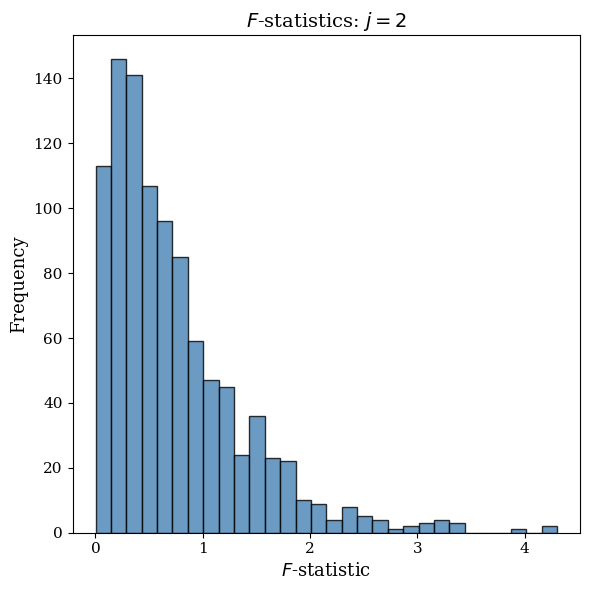}
        \caption{$j=2$}
    \end{subfigure}
    \hfill
    \begin{subfigure}[t]{0.32\textwidth}
        \centering
        \includegraphics[width=\textwidth]{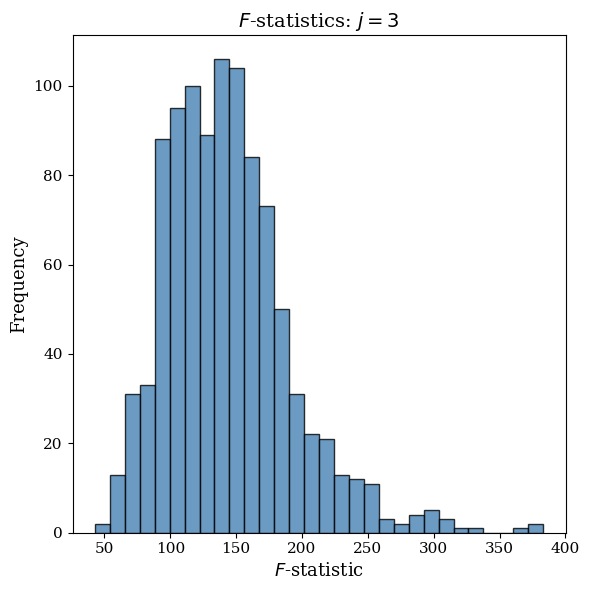}
        \caption{$j=3$}
    \end{subfigure}
    \hfill
    \begin{subfigure}[t]{0.32\textwidth}
        \centering
        \includegraphics[width=\textwidth]{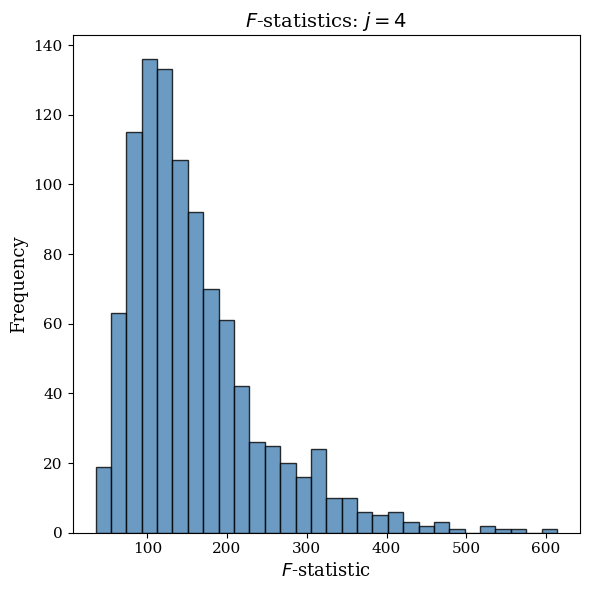}
        \caption{$j=4$}
    \end{subfigure}
    \hfill
    \begin{subfigure}[t]{0.32\textwidth}
        \centering
        \includegraphics[width=\textwidth]{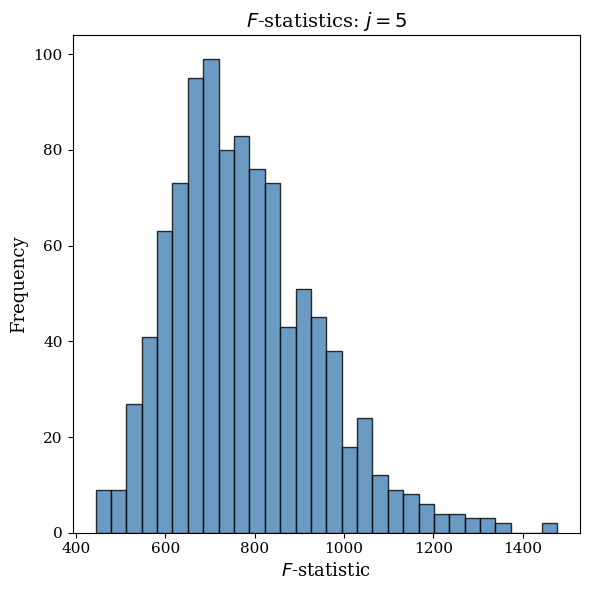}
        \caption{$j=5$}
    \end{subfigure}

    \caption{Distribution of $F$-statistics for different values of $j$.}
    \label{fig:Fs}
\end{figure}

Figure~\ref{fig:Fs} shows the distribution of \( F \)-statistics for \( j = 1, \cdots, 5 \), computed from a sample of size \(1000\) drawn uniformly from a \( d = 3 \) deformed sphere (with parameter \( c = 0.01 \)). A clear shift of the \( F \)-statistic histogram is observed at \( j = 3 \), demonstrating its potential to accurately identify the intrinsic dimension.

In practice, it is often more convenient to base the dimension estimate on the \( p \)-value associated with the \( F \)-statistic. However, this approach requires care. The validity of the \( F \)-distribution hinges on the assumption of normally distributed errors, which may not hold in our setting. When $j = d$, errors in our model (i.e., the $\epsilon$ term in (\ref{eqn-model})) can be attributed to three primary sources: (i) error introduced by estimating the tangent space via PCA, (ii) approximation error using the second-order Taylor polynomial, and (iii) error from noise. None of these is guaranteed to be normal, and therefore the theoretical \( F \)-distribution may not strictly apply.

{\color{black}Despite this, the \(F\)-statistic is known to be robust to various violations of model assumptions; see \citep{box1962robustness,ali1996robustness}. Theorem~\ref{thm:qe-separation} in Section~\ref{sec-analysis} further provides a theoretical guarantee under normal noise: the \(F\)-statistic asymptotically separates the true dimension \(d\) from all lower candidate dimensions. Empirically, we consistently observe a pronounced jump in its value at the true dimension, and applying a fixed \( p \)-value threshold (e.g., at the 1\% level) leads to reliable results across a broad range of datasets and settings.
}

An alternative goodness-of-fit measure that penalizes model complexity is the \emph{adjusted} \( R^2 \) score. This statistic is bounded above by 1, with larger values indicating better fit. A negative adjusted \( R^2 \) suggests that the model performs worse than the trivial model.

The complete \textbf{QE} procedure is summarized in Algorithm~\ref{alg:qe}. Beyond the core framework, we incorporate the following refinements:

\begin{itemize} 
\item Although the first-order Taylor expansion of \( \bm{g} \) is theoretically zero, which implies that our models should only contain second-order terms. In practice, the PCA coordinates only approximate the theoretical new coordinates in Eq.~\eqref{eqn-lgr}. Therefore, we include both first- and second-order terms in the models to better account for the deviation.

\item Since \( p - 1 \) models are fitted in each neighborhood, the procedure can become inefficient when \( p \gg d \). In addition to implementation-level optimizations such as parallelization across neighborhoods, we limit the number of models per neighborhood (i.e., how many candidate dimensions) to a maximum of \( p_{\max} - 1\). The threshold $p_{\max}$ is determined by three factors: the ambient dimension \( p \); the largest index for which the standard deviation of the new coordinates is sufficiently large (to avoid fitting near-constant responses); and the neighborhood size \( K \), as regression becomes unreliable when \( K \) is smaller than the number of predictors. Letting \( q \) denote the largest nonconstant coordinate index, we define\footnote{When $j = p_{\max} - 1$ components are used as inputs, we need $(p_{\max} - 1)(p_{\max} - 2)/2 + 2(p_{\max} - 1) + 1 \leq K + 1$, which leads to the expression in Eq.~\eqref{eqn-pmax}.}: 

\begin{equation} \label{eqn-pmax}
p_{\max} = \min\left\{p,\, q,\, \left\lfloor \frac{\sqrt{9 + 8K} + 1}{2} \right\rfloor \right\}.
\end{equation} 

\item Each neighborhood yields a local estimate \( \hat{d}_k \), which is the smallest index \( j \) such that the \( p \)-value of the corresponding \( F \)-statistic falls below 1\%. Since if $x_{j + 1}'$ is related to $x_{1:j}'$, so does any component after it. To improve robustness, we further require that all models for indices greater than \( j \) also yield \( p \)-values below the same threshold. \textcolor{black}{Note that a $1\%$ threshold implies that the probability of falsely declaring a model significant is controlled at a low level for each individual $F$-test, and the $p$-values from different $F$-tests are positively related since most of the data are reused across regressions. This additional requirement helps keep the probability of making an incorrect estimate low.}

\item To compare models across neighborhoods, we record the adjusted \( R^2 \) of the model corresponding to each \( \hat{d}_k \). This serves as a joint measure of fit and complexity, and should be close to 1 for a well-fitting model. A local estimate is considered valid only if its adjusted \( R^2 \) is positive. The final global estimate \( \hat{d} \) is then computed as the weighted average of all valid local estimates \( \hat{d}_k \), with weights proportional to their adjusted \( R^2 \) scores. \end{itemize}

Compared to ordinary least squares (OLS), the statistical properties of total least squares (TLS) are less well understood. In particular, TLS does not naturally support classical inference tools such as the \( F \)-test or adjusted \( R^2 \), making hypothesis testing and model comparison more challenging. In light of this, we  can directly evaluate model fit using the total error instead, denoted by \( \sigma^2_{j} \). This quantity corresponds to the sum of squared orthogonal distances from the data points to the best-fit affine subspace.

\begin{algorithm}[H]
\caption{\textbf{QE}}
\label{alg:qe}
\begin{algorithmic}[1]
\State \textbf{Input:} sample \( \{\bm{x}_k\}_{k=1}^n \); neighborhood size \( K \)
\State \textbf{Output:} estimated dimension \( \hat{d} \)
\For{\( k = 1 \) to \( n \)}
    \State Find the \( K \)-nearest neighbors of \( \bm{x}_k \)
    \State Perform PCA on \( \bm{x}_k \) and its neighbors; express them in new coordinates as \( \bm{x}_0', \dots, \bm{x}_K' \)
    \State Determine \( p_{\max} \) according to equation~\eqref{eqn-pmax}
    \For{\( j = 1 \) to \( p_{\max} \)}
        \State Fit a quadratic regression model (via OLS) between the first \( j \) coordinates and the \( (j + 1) \)-th
        \State Compute the \( F \)-statistic \( F_j \), its \( p \)-value \( p_j \), and the adjusted \( R^2 \) score \( w_j \)
    \EndFor
    \State Initialize \( \hat{d}_k = p_{\max} \), \( w_k^* = 0 \) for \(k = 1, \cdots, n\)
    \For{\( j = 1 \) to \( p_{\max} \)}
        \If{\( w_j > 0 \) and \( p_i < 1\% \) for all \( i \geq j \)}
            \State Set \( \hat{d}_k = j \), \( w_k^* = w_j \)
            \State \textbf{break}
        \EndIf
    \EndFor
\EndFor
\State \Return \( \hat{d} = \sum_k w_k^* \hat{d}_k \big/ \sum_k w_k^* \) (or the unweighted average if all \( w_k^* = 0 \))
\end{algorithmic}
\end{algorithm}

\begin{figure}[H]
    \centering
    \begin{subfigure}[b]{0.4\textwidth}
        \centering
        \includegraphics[width=\textwidth]{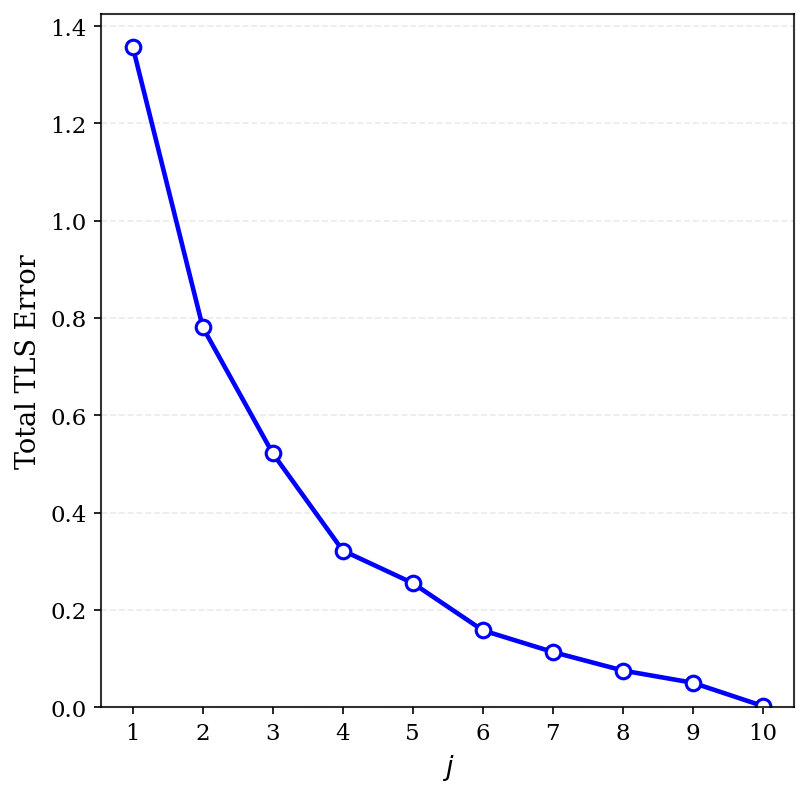}
        \caption{Total TLS error \( \sigma_{j} \).}
        \label{fig:totalerror}
    \end{subfigure}
    \hfill
    \begin{subfigure}[b]{0.4\textwidth}
        \centering
        \includegraphics[width=\textwidth]{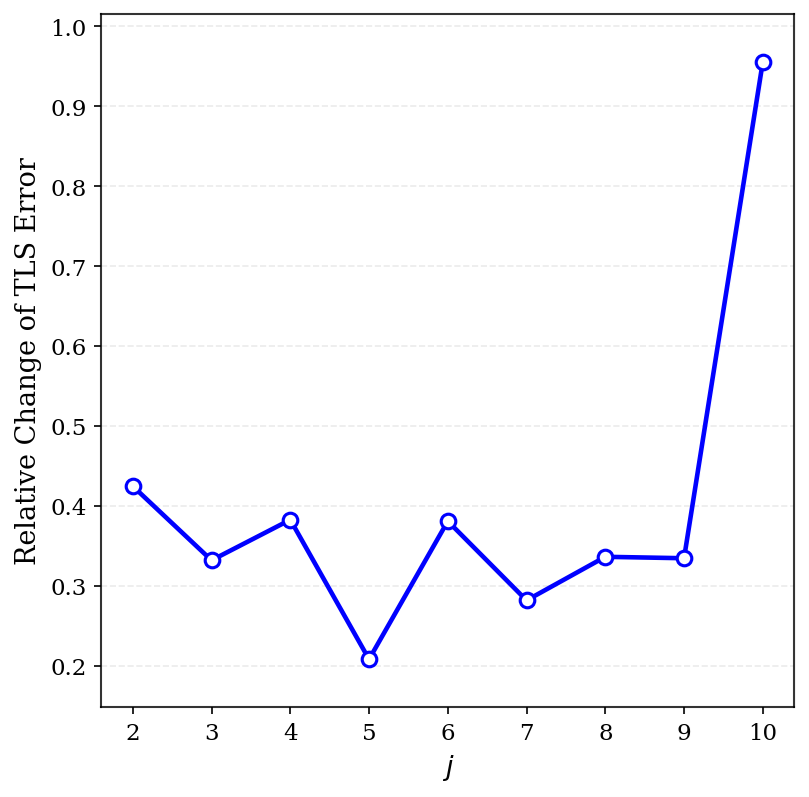}
        \caption{Relative drop in TLS error \( \eta_j \).}
        \label{fig:drops}
    \end{subfigure}
    \caption{Total error and its relative drops as a function of model dimension.}
    \label{fig:rss}
\end{figure}

As in \textbf{QE}, the quadratic model is expected to become appropriate only when \( j \geq d \). While the total TLS error generally decreases as \( j \) increases, a more pronounced drop should be observed at the intrinsic dimension \( d \). However, we found this drop is not always sharp or immediately discernible. Figure~\ref{fig:rss} illustrates this behavior for models with \( j = 1, \dots, 10 \), applied to data sampled from a \( d = 10 \) sphere. Although the error reaches its minimum at \( j = 10 \), the decrease from \( j = 9 \) to \( j = 10 \) is relatively modest.

To improve sensitivity in identifying the intrinsic dimension, we consider the \emph{relative drop} in total error, rather than its absolute value. Specifically, the relative drop from dimension \( j \) to \( j + 1 \) is defined as
\begin{equation} \label{eqn-drop}
    \eta_{j + 1} = \frac{\sigma^2_{j} - \sigma^2_{j + 1}}{\sigma^2_{j}},
\end{equation}
where \( \sigma^2_{j} \) denotes the TLS error associated with dimension \( j \).

As shown in Figure~\ref{fig:drops}, the relative drop \( \eta_j \) peaks at \( j = d \), providing a natural criterion for estimating the local intrinsic dimension \( \hat{d}_k \). Empirically, we find that this approach works much better than the total error measurement. However, it is unable to distinguish the case \( d = 1 \) by design, as there is no $j = 0$ model to compare the total error with. \textcolor{black}{As a result, our \textbf{TLS} estimation is bounded below by $2$, as can be seen from its estimates on the helix dataset ($M_6$, a 1-dimensional manifold) in Table~\ref{tab:res01}.}

The complete \textbf{TLS} procedure is summarized in Algorithm~\ref{alg:tls}.

\begin{algorithm}[H]
\caption{\textbf{TLS}}
\label{alg:tls}
\begin{algorithmic}[1]
\State \textbf{Input:} sample \( \{\bm{x}_k\}_{k=1}^n \); neighborhood size \( K \)
\State \textbf{Output:} estimated dimension \( \hat{d} \)
\For{\( k = 1 \) to \( n \)}
    \State Find the \( K \)-nearest neighbors of \( \bm{x}_k \)
    \State Perform PCA on \( \bm{x}_k \) and its neighbors; express them in new coordinates as \( \bm{x}_0', \dots, \bm{x}_K' \)
    \State Determine \( p_{\max} \) according to Eq.~\eqref{eqn-pmax}
    \For{\( j = 1 \) to \( p_{\max} \)}
        \State Fit a quadratic regression model (via TLS) between the first \( j \) coordinates and the \( (j + 1) \)-th
        \State Compute the total error \( \sigma_{j}^2 \)
    \EndFor
    \For{\( j = 2 \) to \( p_{\max} \)}
        \State Compute the relative drop \( \eta_j \) using Eq.~\eqref{eqn-drop}
    \EndFor
    \State Set \( \hat{d}_k = \arg\max_j \{ \eta_j \} \)
\EndFor
\State \Return \( \hat{d} = \frac{1}{n} \sum_k \hat{d}_k \)
\end{algorithmic}
\end{algorithm}

{\color{black}
\subsection{Properties and Complexity Analysis}\label{sec-analysis}

Manifold dimension estimators aim to estimate the parameter of a distribution supported on a $d$-dimensional manifold embedded in $\mathbb{R}^p$. The flexibility of the underlying geometry, together with the fact that the manifold is a measure-zero set, poses substantial challenges for establishing their theoretical properties. For our \textbf{QE} and \textbf{TLS} estimators, rather than pursuing a complete global consistency theory, we focus on a local neighborhood around any $\bm{x}_0 \in M$ and establish their asymptotic properties under an approximate data-generating model for the $K$ neighboring data points $\bm{x}_0^1, \cdots, \bm{x}_0^K$ around it. See Appendix H for details, which provides detailed descriptions and proofs for the following results.

\begin{proposition}[Population PCA recovers the tangent space]\label{prop:population-pca} Under the local model and some regular condition, the first $d$ principal directions of the population PCA spans the tangent space $T_{\bm{x}_0}M$.
\end{proposition}

\begin{proposition}[The first normal component captures curvature]
\label{prop:first-normal-component}
Suppose that $M$ has nonzero curvature at $\bm{x}_0$, then the $(d+1)$-st population principal component coodinate contains a nonconstant quadratic function of the first $d$ tangent coordinates.
\end{proposition}

\begin{theorem}[Asymptotic separation of the \textbf{QE} statistic]
\label{thm:qe-separation}
Let $F_j$ denote the $F$-test statistic of the \textbf{QE} estimator for candidate dimension $j$. Then $F_j$ is bounded in probability for $j=1,\ldots,d-1$, so is $F_d/K$. Consequently, as $K\to\infty$, the \textbf{QE} procedure asymptotically distinguishes the true intrinsic dimension $d$ from every lower candidate dimension: for $j<d$, $F_j$ remains bounded, whereas $F_d$ explodes.
\end{theorem}

\vspace{0.25cm}

Following the standard results from numerical linear algebra \citep{golub2013matrix} and optimization literature \citep{arya1998optimal, hastie2009elements} on the procedures employed in Algorithms~\ref{alg:qe} and \ref{alg:tls}, both \textbf{QE} and \textbf{TLS} can be shown to have the same overall computational complexity:
\begin{equation}
  O(nKp_{\max}^5 + n p^3 + nKp^2 + n^2 p).  
\end{equation}

From the complexity analysis, we can see that truncating the number of models considered to $p_{\max} - 1$, as determined by the neighborhood size $K$, can significantly improve the computational efficiency of our methods, particularly when the intrinsic dimension $d$ is much smaller than the ambient dimension $p$, the regime where the manifold hypothesis becomes useful. On the other hand, both \textbf{QE} and \textbf{TLS} suffer when $d$ and $p$ are comparable and $p$ is large, since there will be too many candidate models to consider.

Another limitation closely related to above point is that as $d$ increases, our methods require a rapidly increasing neighborhood size to ensure stable regression, as both first and second-order terms are incorporated in the regression model. Consequently, since the neighborhoods must also remain local enough, this in turn implies that the required sample size must also grow substantially with $d$. While most manifold dimension estimators require larger sample sizes as the intrinsic dimension grows, our methods can be more demanding in this respect due to the use of second-order structure.}

\section{Evaluation}

\subsection{Setup}

In this section, we evaluate the performance of our proposed estimators, \textbf{QE} and \textbf{TLS}, on a diverse collection of synthetic and real-world datasets. For the synthetic datasets, each experiment is repeated $100$ times using randomly generated samples. To assess both bias and variance, we report the mean and standard deviation of the resulting estimates. For comparison, we also include eight additional estimators: \textbf{Local PCA}, \textbf{CA-PCA}, \textbf{TwoNN}, \textbf{DanCo}, \textbf{MADA}, \textbf{MLE}, \textbf{TLE} and \textbf{Wasserstein}. These methods span a diverse range of design principles and are either widely used or identified as top-performing approaches in the existing literature\footnote{Our experiments were conducted using \texttt{Python}. Specifically, we utilized the \texttt{skdim} package implementations for \textbf{Local PCA}, \textbf{TwoNN}, \textbf{DanCo}, \textbf{MADA}, \textbf{MLE}, \textbf{TLE}, while the remaining estimators, as well as the sampling and testing procedures, were implemented by ourselves. The code is available at: \url{https://github.com/loong-bi/manifold-dimension-estimation}.
}.

\begin{figure}[H]
    \centering
    \begin{subfigure}[b]{0.45\textwidth}
        \centering
        \includegraphics[width=\textwidth]{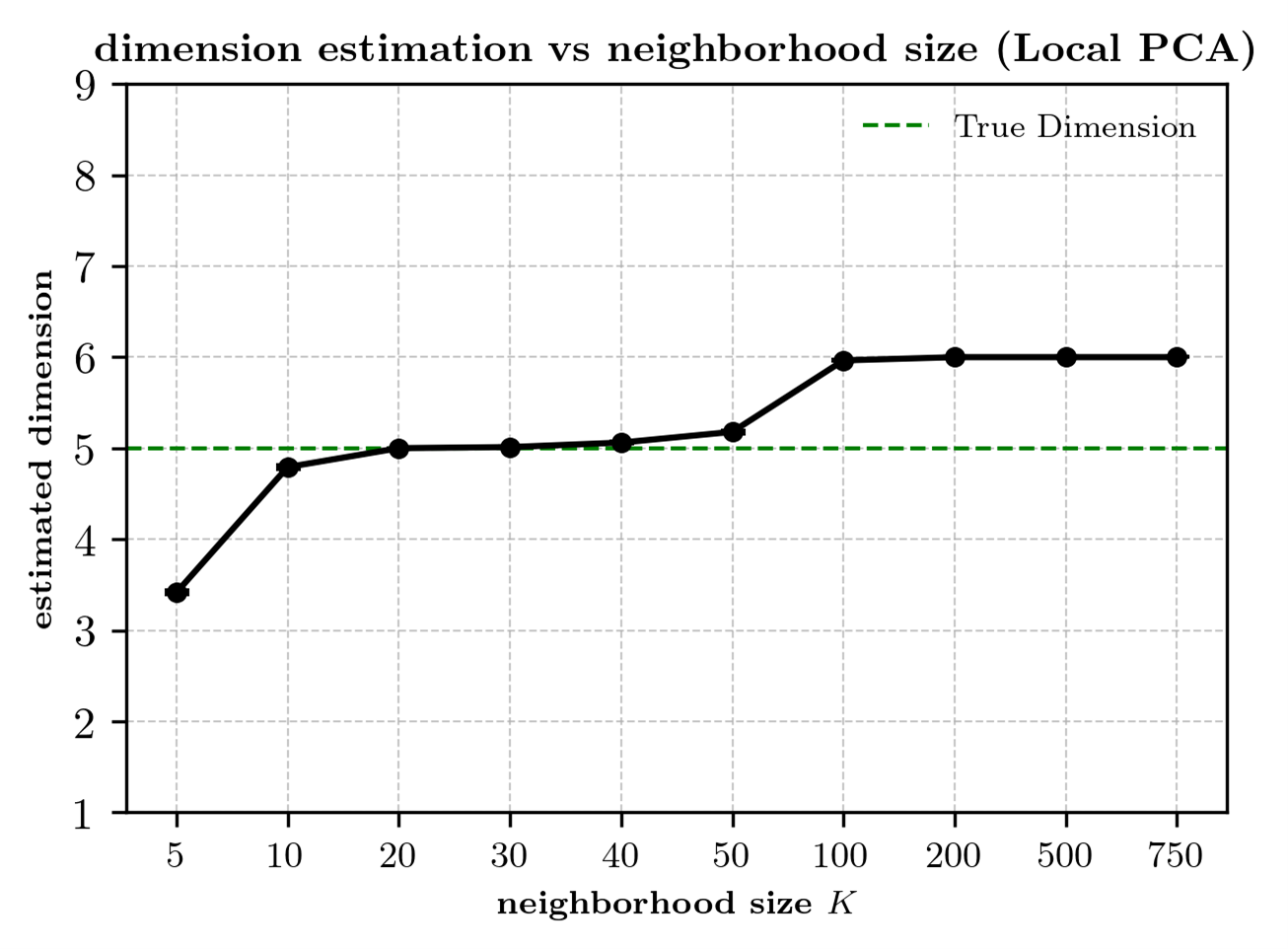}
    \end{subfigure}
    \hfill
    \begin{subfigure}[b]{0.45\textwidth}
        \centering
        \includegraphics[width=\textwidth]{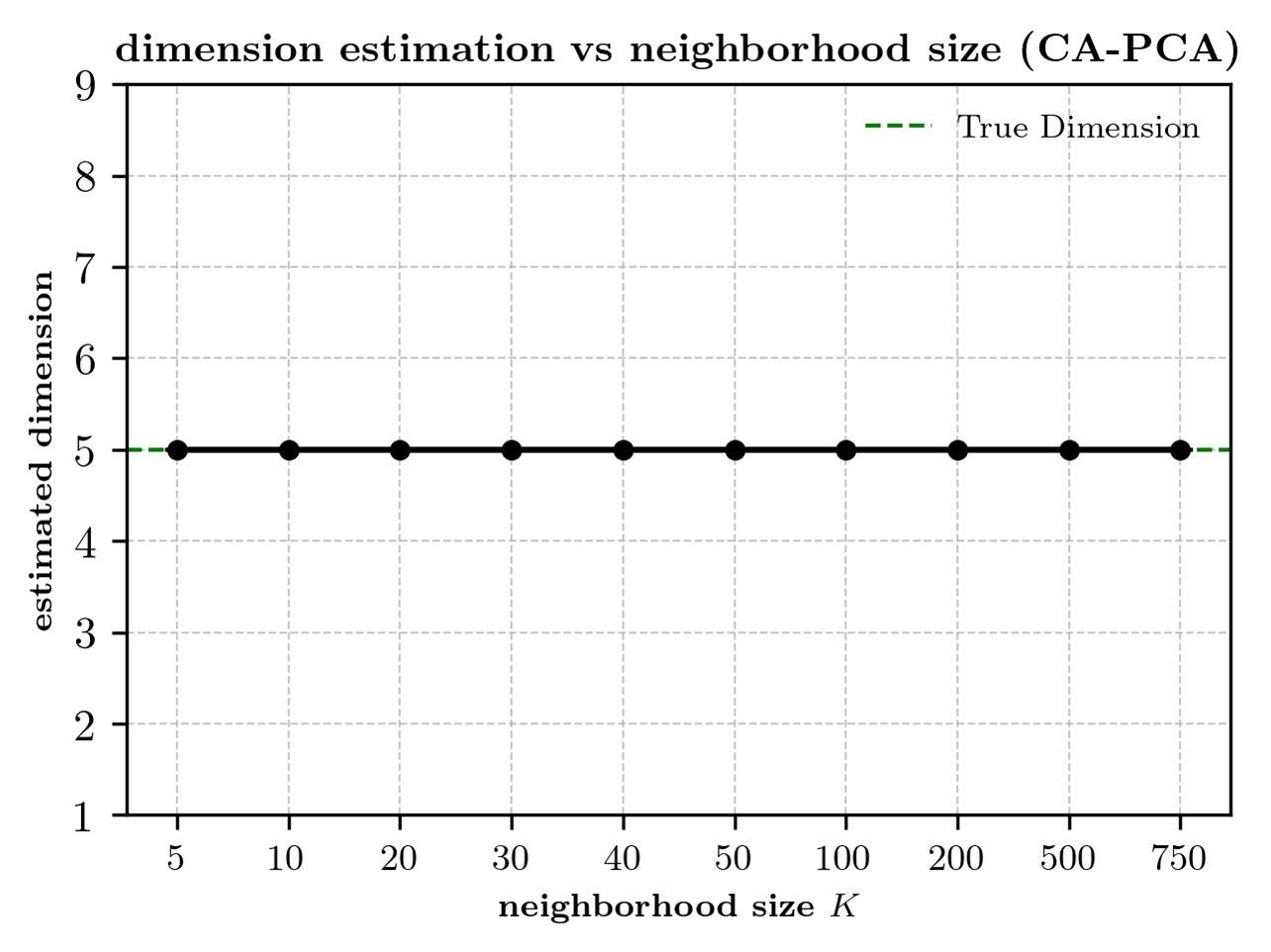}
    \end{subfigure}
    
    \vskip\baselineskip

    \begin{subfigure}[b]{0.45\textwidth}
        \centering
        \includegraphics[width=\textwidth]{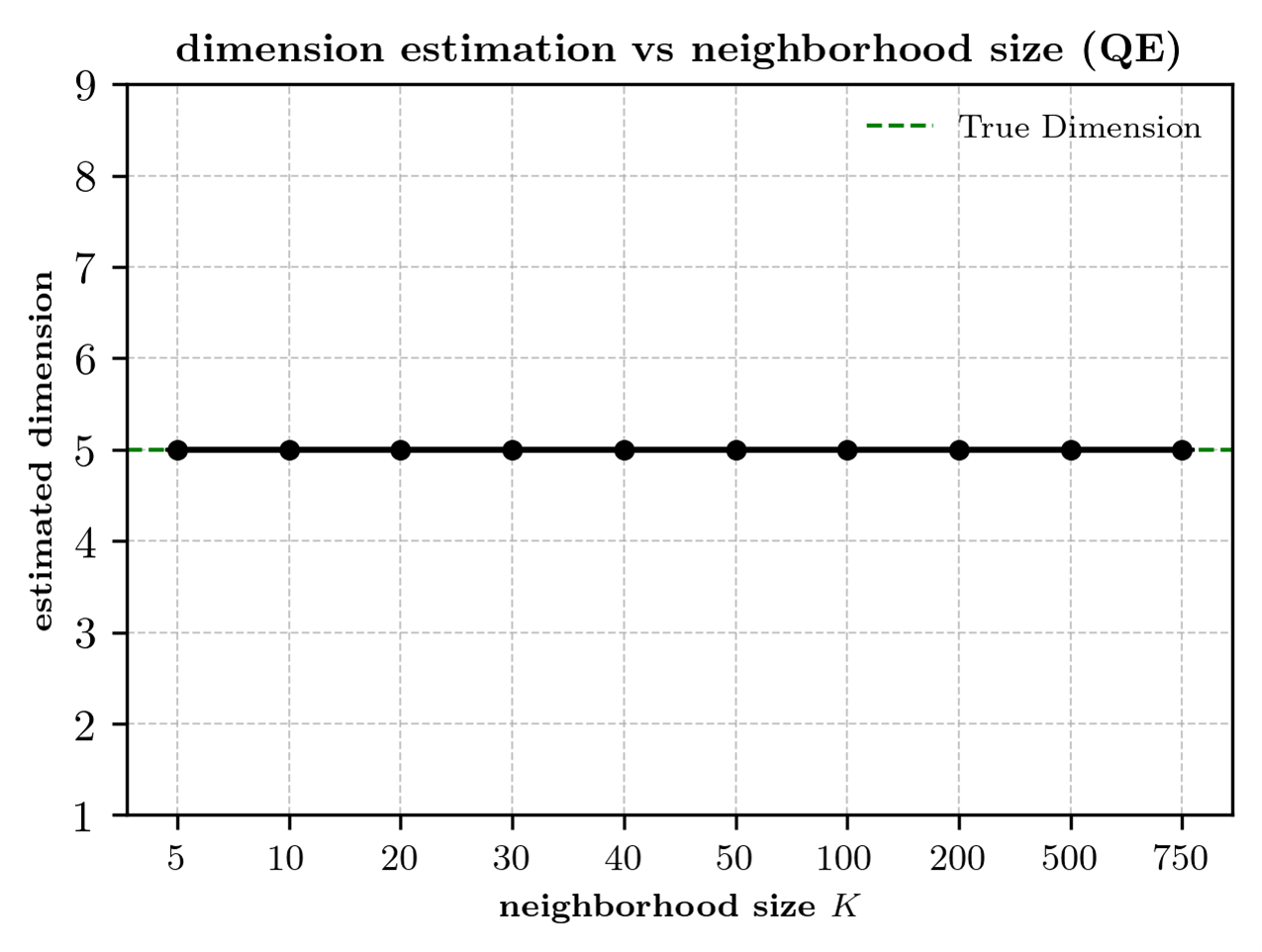}
    \end{subfigure}
    \hfill
    \begin{subfigure}[b]{0.45\textwidth}
        \centering
        \includegraphics[width=\textwidth]{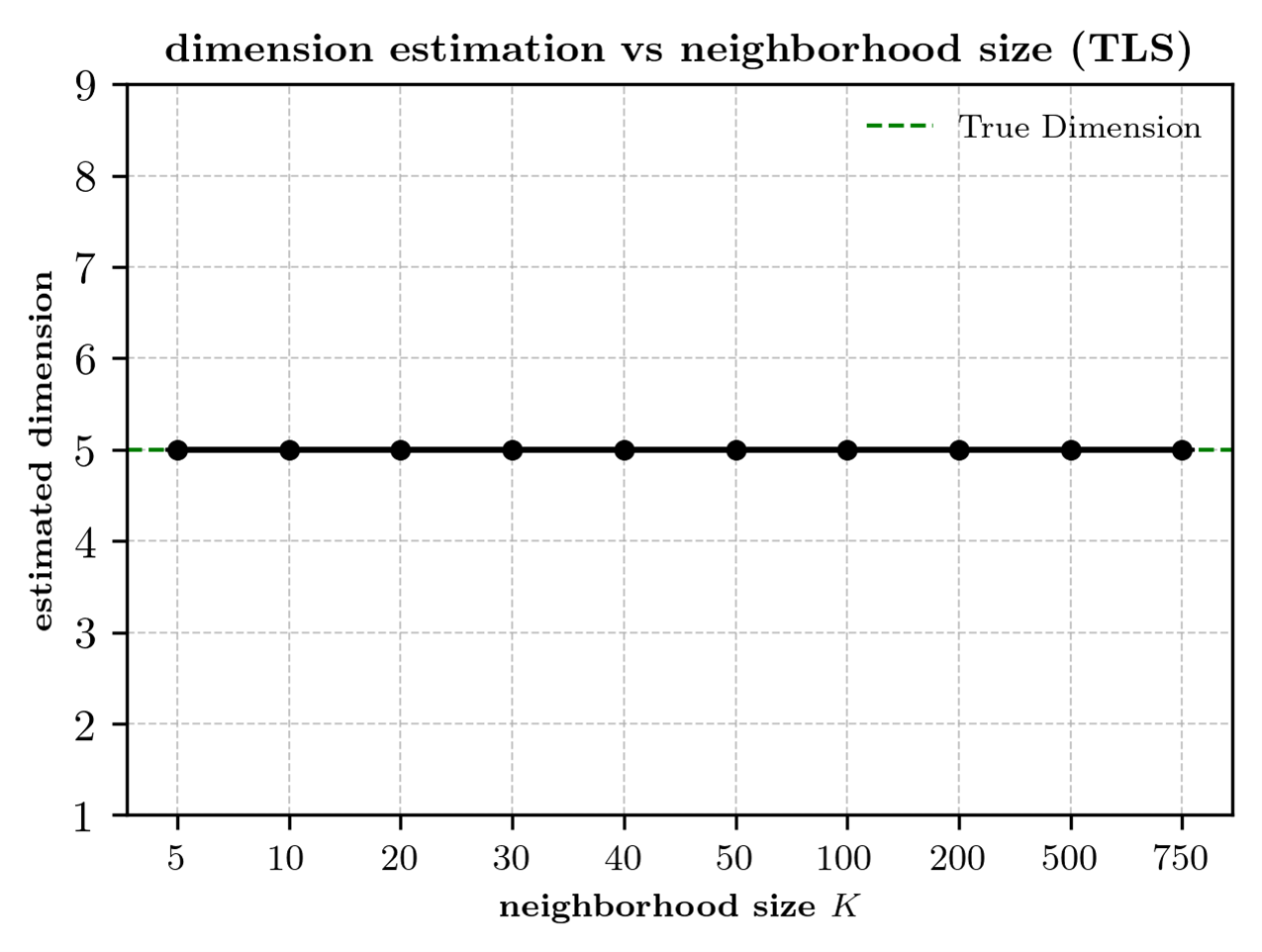}
    \end{subfigure}
    
    \caption{Effect of Neighborhood Size $K$ (top-left: \textbf{Local PCA}; top-right: \textbf{CA-PCA}; bottom-left: \textbf{QE}; bottom-right: \textbf{TLS}).}
    \label{fig:nbh}
\end{figure}

With the exception of \textbf{TwoNN}, all evaluated estimators depend on the neighborhood size $K$ as a hyperparameter. The choice of $K$ has a well-documented impact on performance of flatness-based estimators, as it introduces a trade-off: selecting a small $K$ allows for more accurate local linear approximations due to less spread among neighboring points, but may result in insufficient data to reliably capture the underlying geometry. Conversely, a large $K$ can reduce variance but may lead to biased estimates, as the flatness assumption is violated more. For instance, \textbf{DanCo} typically achieves good performance when $K \doteq 10$. The sensitivity of our proposed estimators, together with \textbf{Local PCA} and \textbf{CA-PCA}, to the choice of $K$ is illustrated in Figure~\ref{fig:nbh}. The results are based on experiments conducted on datasets of size $1000$, uniformly sampled from a 5-dimensional unit sphere embedded linearly in $\mathbb{R}^{10}$.

Several noteworthy patterns can be observed from Figure~\ref{fig:nbh}. As expected for an estimator based on the flatness assumption, the estimates produced by \textbf{Local PCA} tend to increase with neighborhood size. This behavior arises because larger neighborhoods capture more of the global geometry, thereby violating the local linear assumption upon which the estimator relies. In contrast, the estimates from \textbf{CA-PCA} and our proposed methods, \textbf{QE} and \textbf{TLS}, remain remarkably stable across the whole range of neighborhood sizes considered. This robustness sets them apart from many existing methods that are sensitive to the choice of $K$ like \textbf{Local PCA}.

Since \textbf{CA-PCA}, \textbf{QE}, and \textbf{TLS} all explicitly incorporate curvature into the estimation process, their robustness to neighborhood size suggests that doing so enables more effective modeling of local geometry across a broader range of scales. This property is particularly advantageous for our proposed estimators, which, as discussed in Section~\ref{sec-analysis}, require sufficiently large neighborhoods to operate reliably. Consequently, it is both essential and beneficial to evaluate our methods over a wide range of neighborhood sizes.

Finally, since the intrinsic dimension is typically unknown in practice, a fair comparison across estimators requires aggregating results over regions of stable estimates in the neighborhood size $K$. Specifically, we adopt a heuristic approach which is supported by Figure~\ref{fig:nbh}: we search the hyperparameter space for regions in which the standard deviation of the estimates is minimized, i.e., the mean estimates remain approximately constant. If no such stable region is identified, we report the average over all candidate hyperparameter values. For further details, see \citep{bi2025manifold}\footnote{\textcolor{black}{We have also benchmarked the relative runtime of our estimators with respect to changes in intrinsic dimension, ambient dimension, and sample size along with \textbf{TwoNN} and \textbf{CA-PCA}, the results (Table G.1 to G.3 can be found in Appendix G.)}}.

\subsection{Results}

Our first set of experiments evaluates the estimators on datasets supported on 18 manifolds described in Table~\ref{tab:manifolds}. The underlying manifolds for these datasets are commonly studied and identical to those used in our previous survey. Prior evaluations have shown that \textbf{Local PCA}, \textbf{CA-PCA}, \textbf{DanCo}, and \textbf{TwoNN} perform well on these benchmarks. The results of our experiments are summarized in Tables~\ref{tab:res01} to \ref{tab:res02} and Tables D.1 to D.6.

Tables~\ref{tab:res01} and \ref{tab:res02} report results for noiseless, uniformly sampled datasets with sample sizes $n = 500$ and $n = 2000$, respectively. Notably, the selected baseline estimators already perform competitively in this setting. Nonetheless, our proposed estimators demonstrate comparable or superior performance in many cases—particularly when the sample size is small, where competing methods tend to exhibit greater bias and variance, especially on nonlinearly embedded ($M_{41}$ to $M_{43}$) and high-dimensional manifolds ($M_{13}$ and $M_{33}$).

However, our methods encounter difficulties on trivial manifolds such as hyperballs ($M_{21}$ to $M_{23}$). This is expected, as both \textbf{QE} and \textbf{TLS} are designed to fit local quadratic models, which are ill-suited for trivial manifolds with zero curvature, i.e., the quadratic model misspecifies. In these cases, the estimators revert to their fallback mechanisms, which are less effective. 

Results for the same set of synthetic datasets with added $p$-dimensional noise are presented in Tables D.1 to D.6. Three types of noise are considered. Tables D.1 and D.2 report results under Gaussian noise, Tables D.3 and D.4 under uniform noise, and Tables D.5 and D.6 under Laplace noise\footnote{Due to page limitation, Tables D.1 to D.6 are provided in Appendix D.}. The parameters of the noise distributions are chosen so that all noise types have the same variance.

As expected, the presence of noise leads to a substantial degradation in performance across all estimators, with both bias and variance increasing significantly. The type of noise appears to have only a limited impact on the results. Compared to the noiseless setting, the top-performing estimators are more dispersed. Overall, \textbf{QE}, \textbf{TLS}, and \textbf{MADA} perform the best. Our estimators tend to outperform \textbf{MADA} when the sample size is large; however, for smaller sample sizes, \textbf{MADA} appears to be slightly more competitive. In comparison, although \textbf{DanCo}, and \textbf{TwoNN} perform reasonably well on noiseless data, their performance deteriorates a lot in the presence of noise.

One noteworthy observation is that \textbf{QE} and \textbf{TLS} exhibit comparable performance in the presence of noise, even though \textbf{TLS} is, by design, expected to perform better as it accounts for deviations across all components. We conjecture that the $F$-test employed in \textbf{QE} for model evaluation is well grounded and more robust than the heuristic relative error drop criterion used in \textbf{TLS}. With a more reliable evaluation metric, the advantages of \textbf{TLS} may become more evident, which we leave as a direction for future research.

\vspace{0.5cm}

One of the main motivations for explicitly incorporating curvature into the design of manifold dimension estimators is the potential for improved performance on nonlinearly embedded manifolds—i.e., manifolds exhibiting nontrivial curvature in all directions. Such cases are particularly challenging for estimators that rely on local flatness assumptions. The deformed spheres ($M_{41}$ to $M_{43}$) serve as representative examples of this category.

To further evaluate this aspect, we conducted experiments on six additional manifolds generated from the \texttt{skdim} library. These manifolds are all nonlinearly embedded in higher-dimensional ambient spaces and are specifically chosen for their increased geometric complexity. Details of these datasets are provided in Table~\ref{tab:manifolds_skdim}, with the corresponding results presented in Tables~\ref{tab:res09} and \ref{tab:res10} for sample sizes $n = 500$ and $n = 2000$, respectively.

Among these datasets, \textbf{QE} emerges as the top-performing estimator. In contrast, flatness-based methods such as \textbf{DanCo} and \textbf{MLE} exhibit poor performance, which is expected given their reliance on local linearity and lack of explicit curvature modeling. Interestingly, despite accounting for curvature, \textbf{CA-PCA} also performs poorly, especially on manifolds where the ambient dimension significantly exceeds the intrinsic dimension, often leading to substantial overestimation. This phenomenon will be further illustrated on real-world datasets. A possible explanation is that \textbf{CA-PCA} relies, at least in part, on detecting a gap between significant and negligible eigenvalues. When many small but nonzero eigenvalues arise due to curvature or noise, this separation becomes less pronounced, potentially misleading the estimator and degrading its performance.

\vspace{0.5cm}

Lastly, we evaluate our estimators on six real-world datasets. Among them, \texttt{ISOMAP} \citep{tenenbaum2000global}, \texttt{MNIST} and \texttt{ISOLET} \citep{CERUTI20142569} are commonly used benchmarks for manifold dimension estimation, with the underlying manifolds believed to have intrinsic dimensions of $3$, $8$--$11$, and $16$--$22$, respectively. The remaining three datasets, \texttt{Frey Face}, \texttt{Fashion MNIST} \citep{xiao2017fashion}, and \texttt{Gas Sensor Array Drift} \citep{fonollosa2015chemical}, have less consensus regarding their intrinsic dimensionality. The first two are considered more complex than \texttt{ISOMAP} and \texttt{MNIST}, whereas the last is regarded as relatively simple. Applying manifold dimension estimation to these datasets serves as an initial step toward examining the manifold hypothesis. More information about the datasets and the estimation results are presented in Tables~\ref{tab:realdata_ISOMAP} to \ref{tab:realdata_gas}\footnote{The results for \textbf{DanCo}, shown in brackets, are either taken from \citep{CERUTI20142569} and \citep{campadelli2015intrinsic}, which include implementation-specific adjustments designed to stabilize performance, or omitted. A direct application of \textbf{DanCo} using the \texttt{skdim} library does not reproduce these results; see \citep{bi2025manifold} for further discussion.}.

From the results, we observe that \textbf{QE} either produces estimates within the expected range or close to the majority consensus among other estimators, while the estimates of \textbf{TLS} are generally similar to those of \textbf{QE}, albeit slightly lower or higher depending on the dataset. In almost all six datasets, the ambient dimension substantially exceeds the intrinsic dimension, leading to significant overestimation by both \textbf{Local PCA} and \textbf{CA-PCA}, highlighting their limitations in practice. 

Among the remaining methods, \textbf{TwoNN} performs reasonably well, except that it seems to underestimate the intrinsic dimension on \texttt{ISOLET}. For the three datasets without established consensus, the estimators collectively suggest that \texttt{Fashion MNIST} has an intrinsic dimension of approximately $15$--$18$, \texttt{Frey Face} around $6$--$8$, and \texttt{Gas Sensor Air Drift} around $2$--$4$\footnote{\textcolor{black}{We have also measured estimation consistency between our estimators and existing methods on the real-world datasets using pairwise rank correlations. The results are reported in Table~F.1 in Appendix F.}}.

\section{Conclusion}

This article introduces a design paradigm for manifold dimension estimators that seeks to recover the local graph structure of the underlying manifold by performing regression on PCA coordinates. Within this framework, we present two representative estimators: \textbf{QE} and \textbf{TLS}. Both approximate the local geometry using a quadratic model: \textbf{QE} employs ordinary least squares, whereas \textbf{TLS} uses total least squares to account for noise in both predictors and responses. Extensive experiments on synthetic and real-world datasets show that these estimators match or surpass leading state-of-the-art methods, particularly in small-sample settings and in highly nonlinear embedding regimes. 

Our work can be further extended in several directions. First, the role of total least squares in \textbf{TLS} appears to be diminished by its evaluation procedure, and a more refined assessment criterion \textcolor{black}{(e.g., normalize the total error by the degrees of freedom defined specifically for total least squares)} may substantially improve the estimator's performance. Second, both estimators currently rely on PCA for tangent space estimation and coordinate transformations; more robust alternatives, particularly in the presence of noise, \textcolor{black}{such as the proposed method in \citep{aizenbud2025estimation}}, could be explored. Finally, when additional information about the underlying geometry is available, it may be possible to move beyond quadratic models to more expressive formulations. At present, however, the lack of reliable diagnostic tools limits such extensions. Developing suitable sample statistics to address this limitation would therefore be a meaningful direction for future research.

\section*{Acknowledgements}

This research includes computations using the computational cluster Katana supported by Research Technology Services at UNSW Sydney.

\section*{Appendix A: Table of Notation}
\setcounter{table}{0}
\renewcommand{\thetable}{A.\arabic{table}}

\begin{table}[H]
\centering
\caption{Table of Notation.}
\label{tab:notation}
\begin{tabular}{@{} l p{11cm} @{}}
\toprule
\textbf{Notation} & \textbf{Definition} \\
\midrule
$M$         & A $d$-dimensional manifold embedded in $\R^p$. \\
$\{\bm{x}_k\}_{k=1}^n$     & A dataset of size $n$ with elements $\bm{x}_k \in \R^p$. \\
$x_{kj}$     & The $j$-th component of $\bm{x}_k$. \\
$T_{\bm{x}_k}M$          & The tangent space at $\bm{x}_k \in M$. \\
$(T_{\bm{x}_k}M)^\perp$          & The orthogonal complement of $T_{\bm{x}_k}M$. \\
$\bm{x}_{k}^\ell$     & The $\ell$-th nearest neighbor of $\bm{x}_k$. \\
$\bm{x}_\ell'$          & Coordinates of $\bm{x}_{k}^\ell$ with respect to $T_{\bm{x}_k}M$ and $(T_{\bm{x}_k}M)^\perp$. \\
$\bm{g}$          & The local graph function that describes the geometry around $\bm{x}_k \in M$. \\
$g_{\ell}$          & The $\ell$-th component function of $\bm{g}_k$. \\
$\hat{\bm{g}}_j$          & The $j$-dimensional model for estimating $\bm{g}$. \\
$q_{j\ell}$          & The $j$-dimensional quadratic model for estimating $g_{\ell}$. \\
\bottomrule
\end{tabular}
\end{table}

\section*{Appendix B: Table of Manifolds}
\setcounter{table}{0}
\renewcommand{\thetable}{B.\arabic{table}}

\begin{table}[H]
\centering
\caption{The 6 nonlinearly embedded manifolds used in Tables~\ref{tab:res09}--\ref{tab:res10}.}
\label{tab:manifolds_skdim}
\begin{tabular}{@{} l p{11cm} @{}}
\toprule
\textbf{Manifold} & \textbf{Description} \\
\midrule
$M_{\text{NL}1}$         & A $6$-dimensional paraboloid nonlinearly embedded in $\R^{21}$. \\
$M_{\text{NL}2}$     & A $4$-dimensional manifold (concentrated figure) nonlinearly embedded in $\R^{6}$. \\
$M_{\text{NL}3}$          & A $4$-dimensional manifold nonlinearly embedded in $\R^{6}$. \\
$M_{\text{NL}4}$          & A $10$-dimensional manifold nonlinearly embedded in $\R^{40}$. \\
$M_{\text{NL}5}$        & A $12$-dimensional manifold nonlinearly embedded in $\R^{72}$. \\
$M_{\text{NL}6}$          & An $18$-dimensional manifold nonlinearly embedded in $\R^{72}$. \\
\bottomrule
\end{tabular}
\end{table}

\begin{table}[H]
\centering
\caption{The 18 manifolds used in Tables~\ref{tab:res01} to Table D.6.}
\label{tab:manifolds}
\begin{tabular}{@{} l p{11cm} @{}}
\toprule
\textbf{Manifold} & \textbf{Description} \\
\midrule
$M_{11}$         & A $5$-dimensional sphere embedded in $\R^{10}$ with $R = 1$. \\
$M_{12}$     & A $10$-dimensional sphere embedded in $\R^{20}$ with $R = 1$. \\
$M_{13}$          & A $20$-dimensional sphere embedded in $\R^{40}$ with $R = 1$. \\
$M_{21}$          & A $5$-dimensional ball embedded in $\R^{10}$ with $R = 1$. \\
$M_{22}$        & A $10$-dimensional ball embedded in $\R^{20}$ with $R = 1$. \\
$M_{23}$          & A $20$-dimensional ball embedded in $\R^{40}$ with $R = 1$. \\
$M_{31}$           & A $5$-dimensional Gaussian density surface embedded in $\R^{10}$ corresponding to $N(0, 0.25I_5)$. \\
$M_{32}$       & A $10$-dimensional Gaussian density surface embedded in $\R^{20}$ corresponding to $N(0, 0.25I_{10})$. \\
$M_{33}$   & A $20$-dimensional Gaussian density surface embedded in $\R^{40}$ corresponding to $N(0, 0.25I_{20})$. \\
$M_{41}$   & A $3$-dimensional deformed sphere with $c = 0.01$. \\
$M_{42}$         & A $3$-dimensional deformed sphere with $c = 0.1$. \\
$M_{43}$   & A $3$-dimensional deformed sphere with $c = 1$. \\
$M_{5}$     & A $2$-dimensional cylinder embedded in $\R^4$. \\
$M_{6}$   & A $1$-dimensional helix embedded in $\R^3$. \\
$M_{7}$ & A $2$-dimensional Swiss roll embedded in $\R^4$. \\
$M_{8}$      & A $2$-dimensional Mobius strip embedded in $\R^4$. \\
$M_{9}$     & A $2$-dimensional torus embedded in $\R^4$. \\
$M_{10}$         & A $2$-dimensional hyperbolic surface embedded in $\R^4$. \\
\bottomrule
\end{tabular}
\end{table}

\section*{Appendix C: Table of Results}
\setcounter{table}{0}
\renewcommand{\thetable}{C.\arabic{table}}

\begin{landscape}
{\footnotesize
\setlength{\tabcolsep}{2pt}
\begin{longtable}{@{} l| *{10}{c} @{}}
\caption{Mean dimension estimates ($\pm$ standard deviation) for 18 manifolds (Table~\ref{tab:manifolds}) with true dimension $d$, based on 100 replicates of $n=500$ uniform samples. Bold indicates the estimate with minimal MSE.} \label{tab:res01} \\
\toprule
\textbf{Manifold} & \textbf{Local PCA} & \textbf{CA-PCA} & \textbf{DanCo} & \textbf{TwoNN} & \textbf{MLE} & \textbf{TLE} & \textbf{MADA} & \textbf{Wasserstein} & \textbf{QE} & \textbf{TLS} \\
\midrule
\endfirsthead
\multicolumn{11}{c}%
{\tablename\ \thetable\ -- \textit{continued from previous page}} \\
\toprule
\textbf{Manifold} & \textbf{Local PCA} & \textbf{CA-PCA} & \textbf{DanCo} & \textbf{TwoNN} & \textbf{MLE} & \textbf{TLE} & \textbf{MADA} & \textbf{Wasserstein} & \textbf{QE} & \textbf{TLS} \\
\midrule
\endhead
\midrule \multicolumn{9}{r}{\textit{Continued on next page}} \\
\endfoot
\bottomrule
\endlastfoot
$M_{11} (5)$ & $\phantom{1}\bm{5.00 \pm 0.00}$ & $\phantom{1}5.03 \pm 0.01$ & $\phantom{1}6.18 \pm 0.26$ & $\phantom{1}4.90 \pm 0.28$ & $\phantom{1}4.78  \pm  0.04$ & $\phantom{1}5.06  \pm  0.04$ & $\phantom{1}4.65  \pm  0.05$ & $\phantom{1}4.88  \pm  0.08$ & $\phantom{1}\bm{5.00 \pm 0.00}$ & $\phantom{1}\bm{5.00 \pm 0.00}$ \\
$M_{12} (10)$ & $10.21 \pm 0.02$ & $10.10 \pm 0.01$ & $11.16 \pm 0.37$ & $\phantom{1}9.17 \pm 0.53$ & $\phantom{1}8.25  \pm  0.07$ & $\phantom{1}8.48  \pm  0.07$ & $\phantom{1}7.93  \pm  0.10$ & $\phantom{1}8.67  \pm  0.16$ & $\bm{10.00 \pm 0.00}$ & $\bm{10.00 \pm 0.00}$ \\
$M_{13} (20)$ & $20.68 \pm 0.02$ & $19.76 \pm 0.02$ & $20.13 \pm 0.37$ & $15.90 \pm 0.90$ & $13.54 \pm 0.10$ & $13.33 \pm 0.10$ & $12.89 \pm 0.15$ & $14.73 \pm 0.35$ & $\bm{20.00 \pm 0.00}$ & $19.99 \pm 0.00$ \\
$M_{21} (5)$ & $\phantom{1}4.91 \pm 0.03$ & $\phantom{1}\bm{5.00 \pm 0.00}$ & $\phantom{1}4.97 \pm 0.13$ & $\phantom{1}4.64 \pm 0.30$ & $\phantom{1}4.41  \pm  0.04$ & $\phantom{1}4.58  \pm  0.04$ & $\phantom{1}4.24  \pm  0.05$ & $\phantom{1}4.52  \pm  0.07$ & $\phantom{1}4.72 \pm 0.28$ & $\phantom{1}2.86 \pm 0.04$ \\
$M_{22} (10)$ & $\phantom{1}9.93 \pm 0.03$ & $\phantom{1}\bm{9.99 \pm 0.01}$ & $10.03 \pm 0.17$ & $\phantom{1}8.79 \pm 0.51$ & $\phantom{1}7.75  \pm  0.06$ & $\phantom{1}7.88  \pm  0.05$ & $\phantom{1}7.45  \pm  0.08$ & $\phantom{1}8.28  \pm  0.26$ & $\phantom{1}9.84 \pm 0.29$ & $\phantom{1}5.19 \pm 0.18$ \\
$M_{23} (20)$ & $\bm{19.93 \pm 0.02}$ & $19.10 \pm 0.04$ & $20.01 \pm 0.28$ & $15.53 \pm 0.92$ & $13.05 \pm 0.12$ & $12.79 \pm 0.11$ & $12.42 \pm 0.15$ & $14.44 \pm 0.50$ & $19.49 \pm 0.44$ & $18.17 \pm 0.17$ \\
$M_{31} (5)$ & $\phantom{1}5.00 \pm 0.01$ & $\phantom{1}\bm{5.00 \pm 0.00}$ & $\phantom{1}4.97 \pm 0.13$ & $\phantom{1}4.55 \pm 0.27$ & $\phantom{1}4.38  \pm  0.06$ & $\phantom{1}4.54  \pm  0.05$ & $\phantom{1}4.21  \pm  0.06$ & $\phantom{1}4.40  \pm  0.07$ & $\phantom{1}\bm{5.00 \pm 0.00}$ & $\phantom{1}\bm{5.00 \pm 0.00}$ \\
\bottomrule
\end{longtable}}
\end{landscape}

\begin{landscape}
{\footnotesize
\setlength{\tabcolsep}{2pt}
\begin{longtable}{@{} l| *{10}{c} @{}}
\caption{Mean dimension estimates ($\pm$ standard deviation) for 18 manifolds (Table~\ref{tab:manifolds}) with true dimension $d$, based on 100 replicates of $n=2000$ uniform samples. Bold indicates the estimate with minimal MSE.} \label{tab:res02} \\
\toprule
\textbf{Manifold} & \textbf{Local PCA} & \textbf{CA-PCA} & \textbf{DanCo} & \textbf{TwoNN} & \textbf{MLE} & \textbf{TLE} & \textbf{MADA} & \textbf{Wasserstein} & \textbf{QE} & \textbf{TLS} \\
\midrule
\endfirsthead
\multicolumn{11}{c}%
{\tablename\ \thetable\ -- \textit{continued from previous page}} \\
\toprule
\textbf{Manifold} & \textbf{Local PCA} & \textbf{CA-PCA} & \textbf{DanCo} & \textbf{TwoNN} & \textbf{MLE} & \textbf{TLE} & \textbf{MADA} & \textbf{Wasserstein} & \textbf{QE} & \textbf{TLS} \\
\midrule
\endhead
\midrule \multicolumn{9}{r}{\textit{Continued on next page}} \\
\endfoot
\bottomrule
\endlastfoot
$M_{11} (5)$ & $\phantom{1}\bm{5.00 \pm 0.00}$ & $\phantom{1}\bm{5.00 \pm 0.00}$ & $\phantom{1}6.54 \pm 0.17$ & $\phantom{1}4.97 \pm 0.14$ & $\phantom{1}4.92  \pm  0.02$ & $\phantom{1}5.15  \pm  0.02$ & $\phantom{1}4.86  \pm  0.03$ & $\phantom{1}4.89  \pm  0.04$ & $\phantom{1}\bm{5.00 \pm 0.00}$ & $\phantom{1}\bm{5.00 \pm 0.00}$ \\
$M_{12} (10)$ & $\bm{10.00 \pm 0.00}$ & $10.01 \pm 0.00$ & $10.92 \pm 0.14$ & $\phantom{1}9.40 \pm 0.24$ & $\phantom{1}8.89  \pm  0.04$ & $\phantom{1}9.20  \pm  0.04$ & $\phantom{1}8.68  \pm  0.05$ & $\phantom{1}9.06  \pm  0.15$ & $\bm{10.00 \pm 0.00}$ & $\bm{10.00 \pm 0.00}$ \\
$M_{13} (20)$ & $20.17 \pm 0.01$ & $19.70 \pm 0.01$ & $19.97 \pm 0.24$ & $16.74 \pm 0.52$ & $15.12 \pm 0.06$ & $15.13 \pm 0.06$ & $14.61 \pm 0.08$ & $15.52 \pm 0.14$ & $\bm{20.00 \pm 0.00}$ & $19.94 \pm 0.01$ \\
$M_{21} (5)$ & $\phantom{1}4.93 \pm 0.01$ & $\phantom{1}\bm{5.00 \pm 0.00}$ & $\phantom{1}5.00 \pm 0.06$ & $\phantom{1}4.76 \pm 0.14$ & $\phantom{1}4.62  \pm  0.02$ & $\phantom{1}4.76  \pm  0.02$ & $\phantom{1}4.54  \pm  0.03$ & $\phantom{1}4.56  \pm  0.03$ & $\phantom{1}4.31 \pm 0.27$ & $\phantom{1}2.68 \pm 0.02$ \\
$M_{22} (10)$ & $\phantom{1}9.91 \pm 0.02$ & $\phantom{1}\bm{9.99 \pm 0.00}$ & $10.00 \pm 0.11$ & $\phantom{1}8.96 \pm 0.22$ & $\phantom{1}8.41  \pm  0.04$ & $\phantom{1}8.61  \pm  0.04$ & $\phantom{1}8.20  \pm  0.05$ & $\phantom{1}8.45  \pm  0.08$ & $\phantom{1}9.92 \pm 0.21$ & $\phantom{1}4.13 \pm 0.06$ \\
$M_{23} (20)$ & $19.79 \pm 0.03$ & $19.00 \pm 0.03$ & $\bm{19.98 \pm 0.15}$ & $16.17 \pm 0.43$ & $14.59 \pm 0.06$ & $14.51 \pm 0.05$ & $14.09 \pm 0.08$ & $14.94 \pm 0.13$ & $19.45 \pm 0.39$ & $18.18 \pm 0.05$ \\
\bottomrule
\end{longtable}}
\end{landscape}

\begin{landscape}
{\footnotesize
\setlength{\tabcolsep}{2pt}
\begin{longtable}{@{} l| *{10}{c} @{}}
\caption{Mean dimension estimates ($\pm$ standard deviation) for 6 nonlinearly embedded manifolds (Table~\ref{tab:manifolds_skdim}) with true dimension $d$, based on 100 replicates of $n=500$ samples. Bold indicates the estimate with minimal MSE.} \label{tab:res09} \\
\toprule
\textbf{Manifold} & \textbf{Local PCA} & \textbf{CA-PCA} & \textbf{DanCo} & \textbf{TwoNN} & \textbf{MLE} & \textbf{TLE} & \textbf{MADA} & \textbf{Wasserstein} & \textbf{QE} & \textbf{TLS} \\
\midrule
\endfirsthead
\multicolumn{11}{c}%
{\tablename\ \thetable\ -- \textit{continued from previous page}} \\
\toprule
\textbf{Manifold} & \textbf{Local PCA} & \textbf{CA-PCA} & \textbf{DanCo} & \textbf{TwoNN} & \textbf{MLE} & \textbf{TLE} & \textbf{MADA} & \textbf{Wasserstein} & \textbf{QE} & \textbf{TLS} \\
\midrule
\endhead
\midrule \multicolumn{9}{r}{\textit{Continued on next page}} \\
\endfoot
\bottomrule
\endlastfoot
$M_{\text{NL}1} (6)$ & $\phantom{1}7.26 \pm 0.21$ & $\phantom{1}5.68 \pm 0.05$ & $\phantom{1}5.48 \pm 0.16$ & $\phantom{1}5.04 \pm 0.27$ & $\phantom{1}3.61 \pm 0.05$ & $\phantom{1}3.61 \pm 0.06$ & $\phantom{1}3.38 \pm 0.06$ & $\phantom{1}3.67 \pm 0.05$ & $\phantom{1}\bm{5.99 \pm 0.01}$ & $\phantom{1}6.36 \pm 0.03$ \\
$M_{\text{NL}2} (4)$ & $\phantom{1}4.64 \pm 0.09$ & $\phantom{1}5.16 \pm 0.07$ & $\phantom{1}4.90 \pm 0.20$ & $\phantom{1}\bm{3.80 \pm 0.23}$ & $\phantom{1}3.64 \pm 0.05$ & $\phantom{1}3.67 \pm 0.05$ & $\phantom{1}3.56 \pm 0.07$ & $\phantom{1}3.43 \pm 0.04$ & $\phantom{1}4.66 \pm 0.05$ & $\phantom{1}4.56 \pm 0.05$ \\
$M_{\text{NL}3} (4)$ & $\phantom{1}7.65 \pm 0.05$ & $\phantom{1}7.44 \pm 0.06$ & $\phantom{1}6.00 \pm 0.28$ & $\phantom{1}3.91 \pm 0.20$ & $\phantom{1}4.68 \pm 0.07$ & $\phantom{1}4.75 \pm 0.04$ & $\phantom{1}5.07 \pm 0.09$ & $\phantom{1}\bm{4.03 \pm 0.08}$ & $\phantom{1}5.26 \pm 0.10$ & $\phantom{1}6.93 \pm 0.01$ \\
$M_{\text{NL}4} (10)$ & $24.17 \pm 0.84$ & $\phantom{1}\bm{9.08 \pm 0.11}$ & $\phantom{1}6.36 \pm 0.19$ & $\phantom{1}6.11 \pm 0.29$ & $\phantom{1}5.63 \pm 0.14$ & $\phantom{1}4.78 \pm 0.10$ & $\phantom{1}5.47 \pm 0.16$ & $\phantom{1}3.17 \pm 0.05$ & $\phantom{1}9.08 \pm 0.15$ & $10.96 \pm 0.02$ \\
$M_{\text{NL}5} (12)$ & $25.25 \pm 0.36$ & $21.06 \pm 0.08$ & $18.71 \pm 0.79$ & $13.52 \pm 0.72$ & $12.90 \pm 0.11$ & $11.52 \pm 0.12$ & $12.07 \pm 0.13$ & $\phantom{1}5.08 \pm 0.14$ & $\bm{11.98 \pm 0.02}$ & $12.92 \pm 0.01$ \\
$M_{\text{NL}6} (18)$ & $25.52 \pm 0.25$ & $18.04 \pm 0.03$ & $18.28 \pm 0.74$ & $13.68 \pm 0.74$ & $11.89 \pm 0.09$ & $11.25 \pm 0.09$ & $11.29 \pm 0.12$ & $\phantom{1}5.73 \pm 0.10$ & $\bm{18.00 \pm 0.00}$ & $\bm{18.00 \pm 0.00}$ \\
\bottomrule
\end{longtable}

\footnotesize
\setlength{\tabcolsep}{2pt}
\begin{longtable}{@{} l| *{10}{c} @{}}
\caption{Mean dimension estimates ($\pm$ standard deviation) for 6 nonlinearly embedded manifolds (Table~\ref{tab:manifolds_skdim}) with true dimension $d$, based on 100 replicates of $n=2000$ samples. Bold indicates the estimate with minimal MSE.} \label{tab:res10} \\
\toprule
\textbf{Manifold} & \textbf{Local PCA} & \textbf{CA-PCA} & \textbf{DanCo} & \textbf{TwoNN} & \textbf{MLE} & \textbf{TLE} & \textbf{MADA} & \textbf{Wasserstein} & \textbf{QE} & \textbf{TLS} \\
\midrule
\endfirsthead
\multicolumn{11}{c}%
{\tablename\ \thetable\ -- \textit{continued from previous page}} \\
\toprule
\textbf{Manifold} & \textbf{Local PCA} & \textbf{CA-PCA} & \textbf{DanCo} & \textbf{TwoNN} & \textbf{MLE} & \textbf{TLE} & \textbf{MADA} & \textbf{Wasserstein} & \textbf{QE} & \textbf{TLS} \\
\midrule
\endhead
\midrule \multicolumn{9}{r}{\textit{Continued on next page}} \\
\endfoot
\bottomrule
\endlastfoot
$M_{\text{NL}1} (6)$ & $\phantom{1}6.34 \pm 0.06$ & $\phantom{1}5.91 \pm 0.02$ & $\phantom{1}6.61 \pm 0.15$ & $\phantom{1}5.31 \pm 0.13$ & $\phantom{1}4.61 \pm 0.03$ & $\phantom{1}4.63 \pm 0.03$ & $\phantom{1}4.42 \pm 0.03$ & $\phantom{1}4.24 \pm 0.03$ & $\phantom{1}\bm{5.99 \pm 0.00}$ & $\phantom{1}6.36 \pm 0.02$ \\
$M_{\text{NL}2} (4)$ & $\phantom{1}4.44 \pm 0.04$ & $\phantom{1}4.98 \pm 0.03$ & $\phantom{1}5.09 \pm 0.20$ & $\phantom{1}3.86 \pm 0.11$ & $\phantom{1}3.85 \pm 0.02$ & $\phantom{1}\bm{3.92 \pm 0.03}$ & $\phantom{1}3.81 \pm 0.03$ & $\phantom{1}3.66 \pm 0.02$ & $\phantom{1}4.59 \pm 0.02$ & $\phantom{1}4.72 \pm 0.02$ \\
$M_{\text{NL}3} (4)$ & $\phantom{1}4.42 \pm 0.02$ & $\phantom{1}6.28 \pm 0.03$ & $\phantom{1}5.53 \pm 0.18$ & $\phantom{1}3.89 \pm 0.12$ & $\phantom{1}4.24 \pm 0.02$ & $\phantom{1}4.42 \pm 0.02$ & $\phantom{1}4.47 \pm 0.04$ & $\phantom{1}4.20 \pm 0.03$ & $\phantom{1}\bm{4.16 \pm 0.02}$ & $\phantom{1}6.87 \pm 0.01$ \\
$M_{\text{NL}4} (10)$ & $16.07 \pm 0.32$ & $\phantom{1}9.16 \pm 0.04$ & $\phantom{1}7.00 \pm 0.05$ & $\phantom{1}6.45 \pm 0.21$ & $\phantom{1}6.06 \pm 0.05$ & $\phantom{1}5.42 \pm 0.04$ & $\phantom{1}5.93 \pm 0.05$ & $\phantom{1}3.62 \pm 0.03$ & $\phantom{1}\bm{9.34 \pm 0.05}$ & $10.92 \pm 0.01$ \\
$M_{\text{NL}5} (12)$ & $23.92 \pm 0.04$ & $20.33 \pm 0.05$ & $17.36 \pm 0.31$ & $13.43 \pm 0.35$ & $13.63 \pm 0.07$ & $\bm{12.77 \pm 0.06}$ & $13.40 \pm 0.08$ & $\phantom{1}5.80 \pm 0.11$ & $10.98 \pm 0.01$ & $12.92 \pm 0.00$ \\
$M_{\text{NL}6} (18)$ & $22.31 \pm 0.06$ & $17.89 \pm 0.01$ & $20.73 \pm 0.53$ & $14.29 \pm 0.38$ & $13.21 \pm 0.05$ & $12.73 \pm 0.05$ & $12.78 \pm 0.07$ & $\phantom{1}6.73 \pm 0.07$ & $17.00 \pm 0.01$ & $\bm{18.00 \pm 0.00}$ \\
\bottomrule
\end{longtable}}
\end{landscape}

\renewcommand{\arraystretch}{0.8}
\begin{table}[!ht]
\centering
\caption{Dimension Estimate on Real-world Datasets: \texttt{ISOMAP} ($64 \times 64$ sculpture face images)}
\label{tab:realdata_ISOMAP}
\resizebox{\textwidth}{!}{
\begin{tabular}{@{} *{10}{c} @{}}
\toprule
 \textbf{Local PCA} & \textbf{CA-PCA} & \textbf{DanCo} & \textbf{TwoNN} & \textbf{MLE} & \textbf{TLE} & \textbf{MADA} & \textbf{Wasserstein}  & \textbf{QE} & \textbf{TLS} \\
\midrule
 $10.05$ & $25.71$ & ($4.00$) & $3.49$ & $4.34$ & $4.70$ & $5.86$ & $4.13$ & $3.07$ & $3.44$ \\
\bottomrule
\end{tabular}
}
\end{table}

\renewcommand{\arraystretch}{0.8}
\begin{table}[!ht]
\centering
\caption{Dimension Estimate on Real-world Datasets: \texttt{MNIST} ($28 \times 28$ digit images)}
\label{tab:realdata_MNIST}
\resizebox{\textwidth}{!}{
\begin{tabular}{@{} *{10}{c} @{}}
\toprule
 \textbf{Local PCA} & \textbf{CA-PCA} & \textbf{DanCo} & \textbf{TwoNN} & \textbf{MLE} & \textbf{TLE} & \textbf{MADA} & \textbf{Wasserstein} & \textbf{QE} & \textbf{TLS} \\
\midrule
 $15.41$ & $84.66$ & ($9.98$) & $12.98$ & $8.97$ & $7.92$ & $8.67$ & $9.49$ & $9.23$ & $6.73$  \\
\bottomrule
\end{tabular}
}
\end{table}

\renewcommand{\arraystretch}{0.8}
\begin{table}[!ht]
\centering
\caption{Dimension Estimate on Real-world Datasets: \texttt{ISOLET} ($616$-dimensional voice features)}
\label{tab:realdata_ISOLET}
\resizebox{\textwidth}{!}{
\begin{tabular}{@{} *{10}{c} @{}}
\toprule
 \textbf{Local PCA} & \textbf{CA-PCA} & \textbf{DanCo} & \textbf{TwoNN} & \textbf{MLE} & \textbf{TLE} & \textbf{MADA} & \textbf{Wasserstein} & \textbf{QE} & \textbf{TLS} \\
\midrule
 $34.44$ & $54.22$ & ($19.00$) & $9.11$ & $14.29$ & $12.58$ & $13.63$ & $13.38$  & $17.51$ & $25.42$  \\
\bottomrule
\end{tabular}
}
\end{table}

\renewcommand{\arraystretch}{0.8}
\begin{table}[!ht]
\centering
\caption{Dimension Estimate on Real-world Datasets: \texttt{Fashion MNIST} ($28 \times 28$ sneaker images)}
\label{tab:realdata_fashion}
\resizebox{\textwidth}{!}{
\begin{tabular}{@{} *{10}{c} @{}}
\toprule
 \textbf{Local PCA} & \textbf{CA-PCA} & \textbf{DanCo} & \textbf{TwoNN} & \textbf{MLE} & \textbf{TLE} & \textbf{MADA} & \textbf{Wasserstein} & \textbf{QE} & \textbf{TLS} \\
\midrule
 $28.13$ & $54.37$ & - & $16.30$ & $15.63$ & $12.80$ & $14.78$ & $16.58$  & $16.77$ & $17.98$  \\
\bottomrule
\end{tabular}
}
\end{table}

\renewcommand{\arraystretch}{0.8}
\begin{table}[!ht]
\centering
\caption{Dimension Estimate on Real-world Datasets: \texttt{Frey Face} ($20 \times 28$ human face images)}
\label{tab:realdata_frey}
\resizebox{\textwidth}{!}{
\begin{tabular}{@{} *{10}{c} @{}}
\toprule
 \textbf{Local PCA} & \textbf{CA-PCA} & \textbf{DanCo} & \textbf{TwoNN} & \textbf{MLE} & \textbf{TLE} & \textbf{MADA} & \textbf{Wasserstein} & \textbf{QE} & \textbf{TLS} \\
\midrule
 $12.67$ & $54.26$ & - & $5.05$ & $6.84$ & $6.29$ & $7.34$ & $6.20$  & $6.40$ & $10.37$  \\
\bottomrule
\end{tabular}
}
\end{table}

\renewcommand{\arraystretch}{0.8}
\begin{table}[!ht]
\centering
\caption{Dimension Estimate on Real-world Datasets: \texttt{Gas Sensor Array Drift} ($128$-dimensional sensor data)}
\label{tab:realdata_gas}
\resizebox{\textwidth}{!}{
\begin{tabular}{@{} *{10}{c} @{}}
\toprule
 \textbf{Local PCA} & \textbf{CA-PCA} & \textbf{DanCo} & \textbf{TwoNN} & \textbf{MLE} & \textbf{TLE} & \textbf{MADA} & \textbf{Wasserstein} & \textbf{QE} & \textbf{TLS} \\
\midrule
 $3.35$ & $28.16$ & - & $2.87$ & $2.06$ & $1.93$ & $3.02$ & $2.39$  & $4.26$ & $5.32$  \\
\bottomrule
\end{tabular}
}
\end{table}

\section*{Appendix D: Supplementary Table of Results}
\setcounter{table}{0}
\renewcommand{\thetable}{D.\arabic{table}}

\begin{landscape}
{\footnotesize
\setlength{\tabcolsep}{2pt}
\begin{longtable}{@{} l| *{10}{c} @{}}
\caption{Mean dimension estimates ($\pm$ standard deviation) for 18 manifolds (Table B.2) with true dimension $d$, based on 100 replicates of $n=500$ uniform samples perturbed by Gaussian noise. Bold indicates the estimate with minimal MSE.} \label{tab:res03} \\
\toprule
\textbf{Manifold} & \textbf{Local PCA} & \textbf{CA-PCA} & \textbf{DanCo} & \textbf{TwoNN} & \textbf{MLE} & \textbf{TLE} & \textbf{MADA} & \textbf{Wasserstein} & \textbf{QE} & \textbf{TLS} \\
\midrule
\endfirsthead
\multicolumn{11}{c}%
{\tablename\ \thetable\ -- \textit{continued from previous page}} \\
\toprule
\textbf{Manifold} & \textbf{Local PCA} & \textbf{CA-PCA} & \textbf{DanCo} & \textbf{TwoNN} & \textbf{MLE} & \textbf{TLE} & \textbf{MADA} & \textbf{Wasserstein} & \textbf{QE} & \textbf{TLS} \\
\midrule
\endhead
\midrule \multicolumn{9}{r}{\textit{Continued on next page}} \\
\endfoot
\bottomrule
\endlastfoot
$M_{11} (5)$ & $\phantom{1}7.35 \pm 0.32$ & $\phantom{1}7.19 \pm 0.07$ & $\phantom{1}7.49 \pm 0.24$ & $\phantom{1}6.99 \pm 0.35$ & $\phantom{1}5.59  \pm  0.05$ & $\phantom{1}5.80  \pm  0.05$ & $\phantom{1}\bm{5.22  \pm  0.07}$ & $\phantom{1}6.06  \pm  0.11$ & $\phantom{1}8.40 \pm 0.21$ & $\phantom{1}5.55 \pm 0.01$ \\
$M_{12} (10)$ & $19.06 \pm 0.06$ & $13.27 \pm 0.08$ & $14.69 \pm 0.41$ & $11.98 \pm 0.61$ & $\phantom{1}\bm{9.67  \pm  0.08}$ & $\phantom{1}9.61  \pm  0.07$ & $\phantom{1}9.17  \pm  0.10$ & $10.70 \pm 0.23$ & $11.06 \pm 0.15$ & $11.61 \pm 0.01$ \\
$M_{13} (20)$ & $35.93 \pm 0.07$ & $25.73 \pm 0.11$ & $34.36 \pm 1.01$ & $20.84 \pm 1.13$ & $16.37 \pm 0.12$ & $15.32 \pm 0.11$ & $15.41 \pm 0.15$ & $18.90 \pm 0.56$ & $21.05 \pm 0.13$ & $\bm{19.99 \pm 0.00}$ \\
$M_{21} (5)$ & $\phantom{1}9.89 \pm 0.03$ & $\phantom{1}8.77 \pm 0.08$ & $\phantom{1}8.22 \pm 0.22$ & $\phantom{1}7.52 \pm 0.43$ & $\phantom{1}5.65  \pm  0.05$ & $\phantom{1}5.74  \pm  0.05$ & $\phantom{1}\bm{5.19  \pm  0.06}$ & $\phantom{1}6.07  \pm  0.09$ & $\phantom{1}5.77 \pm 0.19$ & $\phantom{1}8.15 \pm 0.01$ \\
$M_{22} (10)$ & $19.50 \pm 0.04$ & $15.20 \pm 0.12$ & $14.65 \pm 0.57$ & $12.25 \pm 0.63$ & $\phantom{1}\bm{9.53  \pm  0.08}$ & $\phantom{1}9.35  \pm  0.07$ & $\phantom{1}8.99  \pm  0.10$ & $10.88 \pm 0.39$ & $10.71 \pm 0.12$ & $11.79 \pm 0.02$ \\
$M_{23} (20)$ & $36.23 \pm 0.08$ & $25.89 \pm 0.12$ & $35.01 \pm 1.25$ & $21.03 \pm 0.97$ & $16.24 \pm 0.13$ & $15.11 \pm 0.12$ & $15.25 \pm 0.16$ & $18.97 \pm 0.76$ & $20.38 \pm 0.12$ & $\bm{19.99 \pm 0.00}$ \\
$M_{31} (5)$ & $\phantom{1}5.06 \pm 0.03$ & $\phantom{1}\bm{5.04 \pm 0.02}$ & $\phantom{1}6.02 \pm 0.15$ & $\phantom{1}5.66 \pm 0.33$ & $\phantom{1}4.74  \pm  0.06$ & $\phantom{1}4.91  \pm  0.06$ & $\phantom{1}4.45  \pm  0.06$ & $\phantom{1}4.80  \pm  0.08$ & $\phantom{1}5.74 \pm 0.15$ & $\phantom{1}6.06 \pm 0.01$ \\
\bottomrule
\end{longtable}}
\end{landscape}

\begin{landscape}
{\footnotesize
\setlength{\tabcolsep}{2pt}
\begin{longtable}{@{} l| *{10}{c} @{}}
\caption{Mean dimension estimates ($\pm$ standard deviation) for 18 manifolds (Table B.2) with true dimension $d$, based on 100 replicates of $n=2000$ uniform samples perturbed by Gaussian noise. Bold indicates the estimate with minimal MSE.} \label{tab:res04} \\
\toprule
\textbf{Manifold} & \textbf{Local PCA} & \textbf{CA-PCA} & \textbf{DanCo} & \textbf{TwoNN} & \textbf{MLE} & \textbf{TLE} & \textbf{MADA} & \textbf{Wasserstein} & \textbf{QE} & \textbf{TLS} \\
\midrule
\endfirsthead
\multicolumn{11}{c}%
{\tablename\ \thetable\ -- \textit{continued from previous page}} \\
\toprule
\textbf{Manifold} & \textbf{Local PCA} & \textbf{CA-PCA} & \textbf{DanCo} & \textbf{TwoNN} & \textbf{MLE} & \textbf{TLE} & \textbf{MADA} & \textbf{Wasserstein} & \textbf{QE} & \textbf{TLS} \\
\midrule
\endhead
\midrule \multicolumn{9}{r}{\textit{Continued on next page}} \\
\endfoot
\bottomrule
\endlastfoot
$M_{11} (5)$ & $\phantom{1}9.78 \pm 0.04$ & $\phantom{1}8.40 \pm 0.03$ & $10.00 \pm 0.02$ & $\phantom{1}7.77 \pm 0.20$ & $\phantom{1}6.22  \pm  0.03$ & $\phantom{1}6.38  \pm  0.03$ & $\phantom{1}5.90  \pm  0.04$ & $\phantom{1}6.53  \pm  0.05$ & $\phantom{1}7.33 \pm 0.01$ & $\phantom{1}\bm{5.52 \pm 0.01}$ \\
$M_{12} (10)$ & $19.60 \pm 0.02$ & $14.56 \pm 0.05$ & $14.89 \pm 0.21$ & $13.03 \pm 0.32$ & $10.98 \pm 0.05$ & $10.96 \pm 0.04$ & $\bm{10.50 \pm 0.05}$ & $11.83 \pm 0.20$ & $13.00 \pm 0.01$ & $10.60 \pm 0.01$ \\
$M_{13} (20)$ & $35.38 \pm 0.04$ & $26.17 \pm 0.05$ & $36.49 \pm 0.90$ & $22.68 \pm 0.64$ & $19.00 \pm 0.08$ & $17.99 \pm 0.07$ & $18.08 \pm 0.09$ & $\bm{20.44 \pm 0.21}$ & $21.00 \pm 0.00$ & $21.00 \pm 0.00$ \\
$M_{21} (5)$ & $\phantom{1}9.99 \pm 0.00$ & $\phantom{1}9.61 \pm 0.02$ & $10.00 \pm 0.00$ & $\phantom{1}8.25 \pm 0.24$ & $\phantom{1}6.53  \pm  0.03$ & $\phantom{1}6.58  \pm  0.03$ & $\phantom{1}6.14  \pm  0.04$ & $\phantom{1}7.06  \pm  0.11$ & $\phantom{1}\bm{5.67 \pm 0.03}$ & $\phantom{1}8.09 \pm 0.03$ \\
$M_{22} (10)$ & $19.83 \pm 0.01$ & $16.21 \pm 0.06$ & $15.02 \pm 0.13$ & $13.33 \pm 0.30$ & $11.01 \pm 0.05$ & $10.84 \pm 0.04$ & $\bm{10.46 \pm 0.06}$ & $11.77 \pm 0.13$ & $11.66 \pm 0.01$ & $12.75 \pm 0.01$ \\
$M_{23} (20)$ & $35.68 \pm 0.04$ & $26.38 \pm 0.06$ & $37.51 \pm 0.96$ & $22.81 \pm 0.58$ & $18.94 \pm 0.08$ & $17.81 \pm 0.08$ & $18.00 \pm 0.10$ & $20.46 \pm 0.20$ & $\bm{20.00 \pm 0.00}$ & $19.99 \pm 0.00$ \\
$M_{31} (5)$ & $\phantom{1}5.08 \pm 0.02$ & $\phantom{1}5.87 \pm 0.05$ & $\phantom{1}6.97 \pm 0.09$ & $\phantom{1}6.34 \pm 0.17$ & $\phantom{1}\bm{5.05  \pm  0.02}$ & $\phantom{1}5.18  \pm  0.02$ & $\phantom{1}4.81  \pm  0.03$ & $\phantom{1}5.27  \pm  0.11$ & $\phantom{1}6.33 \pm 0.01$ & $\phantom{1}6.16 \pm 0.01$ \\
\bottomrule
\end{longtable}}
\end{landscape}

\begin{landscape}
{\footnotesize
\setlength{\tabcolsep}{2pt}
\begin{longtable}{@{} l| *{10}{c} @{}}
\caption{Mean dimension estimates ($\pm$ standard deviation) for 18 manifolds (Table B.2) with true dimension $d$, based on 100 replicates of $n=500$ uniform samples perturbed by uniform noise. Bold indicates the estimate with minimal MSE.} \label{tab:res05} \\
\toprule
\textbf{Manifold} & \textbf{Local PCA} & \textbf{CA-PCA} & \textbf{DanCo} & \textbf{TwoNN} & \textbf{MLE} & \textbf{TLE} & \textbf{MADA} & \textbf{Wasserstein} & \textbf{QE} & \textbf{TLS} \\
\midrule
\endfirsthead
\multicolumn{11}{c}%
{\tablename\ \thetable\ -- \textit{continued from previous page}} \\
\toprule
\textbf{Manifold} & \textbf{Local PCA} & \textbf{CA-PCA} & \textbf{DanCo} & \textbf{TwoNN} & \textbf{MLE} & \textbf{TLE} & \textbf{MADA} & \textbf{Wasserstein} & \textbf{QE} & \textbf{TLS} \\
\midrule
\endhead
\midrule \multicolumn{9}{r}{\textit{Continued on next page}} \\
\endfoot
\bottomrule
\endlastfoot
$M_{11} (5)$ & $\phantom{1}7.20 \pm 0.07$ & $\phantom{1}7.29 \pm 0.33$ & $\phantom{1}7.83 \pm 0.20$ & $\phantom{1}7.06 \pm 0.42$ & $\phantom{1}5.57 \pm 0.04$ & $\phantom{1}5.79 \pm 0.04$ & $\phantom{1}\bm{5.21 \pm 0.06}$ & $\phantom{1}5.74 \pm 0.03$ & $\phantom{1}8.52 \pm 0.35$ & $\phantom{1}5.55 \pm 0.01$ \\
$M_{12} (10)$ & $13.26 \pm 0.08$ & $19.09 \pm 0.06$ & $14.10 \pm 0.27$ & $12.07 \pm 0.74$ & $\phantom{1}\bm{9.66 \pm 0.08}$ & $\phantom{1}9.61 \pm 0.08$ & $\phantom{1}9.15 \pm 0.11$ & $\phantom{1}6.83 \pm 0.04$ & $11.06 \pm 0.15$ & $11.61 \pm 0.01$ \\
$M_{13} (20)$ & $25.75 \pm 0.10$ & $35.87 \pm 0.07$ & $33.71 \pm 0.96$ & $21.18 \pm 1.15$ & $16.36 \pm 0.11$ & $15.36 \pm 0.10$ & $15.37 \pm 0.16$ & $\phantom{1}5.99 \pm 0.05$ & $20.95 \pm 0.15$ & $\bm{19.99 \pm 0.00}$ \\
$M_{21} (5)$ & $\phantom{1}8.81 \pm 0.09$ & $\phantom{1}9.91 \pm 0.03$ & $\phantom{1}7.95 \pm 0.11$ & $\phantom{1}7.45 \pm 0.37$ & $\phantom{1}5.64 \pm 0.06$ & $\phantom{1}5.74 \pm 0.05$ & $\phantom{1}\bm{5.18 \pm 0.06}$ & $\phantom{1}5.22 \pm 0.04$ & $\phantom{1}5.75 \pm 0.15$ & $\phantom{1}8.13 \pm 0.01$ \\
$M_{22} (10)$ & $15.26 \pm 0.13$ & $19.53 \pm 0.04$ & $14.78 \pm 0.72$ & $12.27 \pm 0.56$ & $\phantom{1}\bm{9.53 \pm 0.08}$ & $\phantom{1}9.36 \pm 0.06$ & $\phantom{1}8.98 \pm 0.09$ & $\phantom{1}6.34 \pm 0.09$ & $10.75 \pm 0.19$ & $11.77 \pm 0.02$ \\
$M_{23} (20)$ & $25.92 \pm 0.10$ & $36.17 \pm 0.07$ & $34.58 \pm 1.24$ & $21.21 \pm 0.89$ & $16.23 \pm 0.12$ & $15.15 \pm 0.11$ & $15.24 \pm 0.17$ & $\phantom{1}5.82 \pm 0.06$ & $20.39 \pm 0.15$ & $\bm{19.99 \pm 0.00}$ \\
$M_{31} (5)$ & $\phantom{1}\bm{5.02 \pm 0.01}$ & $\phantom{1}5.05 \pm 0.02$ & $\phantom{1}6.03 \pm 0.15$ & $\phantom{1}5.59 \pm 0.29$ & $\phantom{1}4.73 \pm 0.06$ & $\phantom{1}4.90 \pm 0.05$ & $\phantom{1}4.44 \pm 0.07$ & $\phantom{1}4.50 \pm 0.03$ & $\phantom{1}6.78 \pm 0.17$ & $\phantom{1}6.04 \pm 0.01$ \\
\bottomrule
\end{longtable}}
\end{landscape}

\begin{landscape}
{\footnotesize
\setlength{\tabcolsep}{2pt}
\begin{longtable}{@{} l| *{10}{c} @{}}
\caption{Mean dimension estimates ($\pm$ standard deviation) for 18 manifolds (Table B.2) with true dimension $d$, based on 100 replicates of $n=2000$ uniform samples perturbed by uniform noise. Bold indicates the estimate with minimal MSE.} \label{tab:res06} \\
\toprule
\textbf{Manifold} & \textbf{Local PCA} & \textbf{CA-PCA} & \textbf{DanCo} & \textbf{TwoNN} & \textbf{MLE} & \textbf{TLE} & \textbf{MADA} & \textbf{Wasserstein} & \textbf{QE} & \textbf{TLS} \\
\midrule
\endfirsthead
\multicolumn{11}{c}%
{\tablename\ \thetable\ -- \textit{continued from previous page}} \\
\toprule
\textbf{Manifold} & \textbf{Local PCA} & \textbf{CA-PCA} & \textbf{DanCo} & \textbf{TwoNN} & \textbf{MLE} & \textbf{TLE} & \textbf{MADA} & \textbf{Wasserstein} & \textbf{QE} & \textbf{TLS} \\
\midrule
\endhead
\midrule \multicolumn{9}{r}{\textit{Continued on next page}} \\
\endfoot
\bottomrule
\endlastfoot
$M_{11} (5)$ & $\phantom{1}8.46 \pm 0.03$ & $\phantom{1}9.83 \pm 0.03$ & $\phantom{1}8.13 \pm 0.13$ & $\phantom{1}7.79 \pm 0.20$ & $\phantom{1}6.20 \pm 0.03$ & $\phantom{1}6.38 \pm 0.02$ & $\phantom{1}5.88 \pm 0.03$ & $\phantom{1}6.11 \pm 0.02$ & $\phantom{1}7.33 \pm 0.01$ & $\phantom{1}\bm{5.52 \pm 0.01}$ \\
$M_{12} (10)$ & $14.60 \pm 0.05$ & $19.62 \pm 0.02$ & $15.12 \pm 0.28$ & $13.00 \pm 0.36$ & $10.96 \pm 0.05$ & $10.96 \pm 0.05$ & $\bm{10.47 \pm 0.06}$ & $\phantom{1}7.81 \pm 0.03$ & $13.00 \pm 0.01$ & $10.60 \pm 0.01$ \\
$M_{13} (20)$ & $26.22 \pm 0.05$ & $35.34 \pm 0.04$ & $35.94 \pm 0.77$ & $22.70 \pm 0.61$ & $18.98 \pm 0.08$ & $18.04 \pm 0.07$ & $18.09 \pm 0.10$ & $\phantom{1}6.91 \pm 0.04$ & $\bm{21.00 \pm 0.00}$ & $\bm{21.00 \pm 0.00}$ \\
$M_{21} (5)$ & $\phantom{1}9.67 \pm 0.02$ & $\phantom{1}10.00 \pm 0.00$ & $\phantom{1}9.46 \pm 0.15$ & $\phantom{1}8.25 \pm 0.20$ & $\phantom{1}6.53 \pm 0.03$ & $\phantom{1}6.60 \pm 0.03$ & $\phantom{1}6.13 \pm 0.04$ & $\phantom{1}5.81 \pm 0.02$ & $\phantom{1}\bm{5.67 \pm 0.01}$ & $\phantom{1}8.02 \pm 0.03$ \\
$M_{22} (10)$ & $16.29 \pm 0.05$ & $19.85 \pm 0.01$ & $15.04 \pm 0.16$ & $13.40 \pm 0.34$ & $10.99 \pm 0.05$ & $10.85 \pm 0.04$ & $\bm{10.44 \pm 0.06}$ & $\phantom{1}7.19 \pm 0.05$ & $11.66 \pm 0.01$ & $12.74 \pm 0.01$ \\
$M_{23} (20)$ & $26.45 \pm 0.06$ & $35.64 \pm 0.04$ & $37.75 \pm 0.84$ & $22.84 \pm 0.59$ & $18.95 \pm 0.07$ & $17.89 \pm 0.07$ & $18.01 \pm 0.09$ & $\phantom{1}6.66 \pm 0.05$ & $\bm{20.00 \pm 0.00}$ & $19.99 \pm 0.00$ \\
$M_{31} (5)$ & $\phantom{1}5.78 \pm 0.04$ & $\phantom{1}5.07 \pm 0.01$ & $\phantom{1}7.29 \pm 0.16$ & $\phantom{1}6.37 \pm 0.16$ & $\phantom{1}\bm{5.03 \pm 0.02}$ & $\phantom{1}5.17 \pm 0.02$ & $\phantom{1}4.80 \pm 0.03$ & $\phantom{1}4.94 \pm 0.01$ & $\phantom{1}6.33 \pm 0.01$ & $\phantom{1}6.13 \pm 0.01$ \\
\bottomrule
\end{longtable}}
\end{landscape}

\begin{landscape}
{\footnotesize
\setlength{\tabcolsep}{2pt}
\begin{longtable}{@{} l| *{10}{c} @{}}
\caption{Mean dimension estimates ($\pm$ standard deviation) for 18 manifolds (Table B.2) with true dimension $d$, based on 100 replicates of $n=500$ uniform samples perturbed by Laplace noise. Bold indicates the estimate with minimal MSE.} \label{tab:res07} \\
\toprule
\textbf{Manifold} & \textbf{Local PCA} & \textbf{CA-PCA} & \textbf{DanCo} & \textbf{TwoNN} & \textbf{MLE} & \textbf{TLE} & \textbf{MADA} & \textbf{Wasserstein} & \textbf{QE} & \textbf{TLS} \\
\midrule
\endfirsthead
\multicolumn{11}{c}%
{\tablename\ \thetable\ -- \textit{continued from previous page}} \\
\toprule
\textbf{Manifold} & \textbf{Local PCA} & \textbf{CA-PCA} & \textbf{DanCo} & \textbf{TwoNN} & \textbf{MLE} & \textbf{TLE} & \textbf{MADA} & \textbf{Wasserstein} & \textbf{QE} & \textbf{TLS} \\
\midrule
\endhead
\midrule \multicolumn{9}{r}{\textit{Continued on next page}} \\
\endfoot
\bottomrule
\endlastfoot
$M_{11} (5)$ & $\phantom{1}7.00 \pm 0.08$ & $\phantom{1}6.85 \pm 0.31$ & $\phantom{1}7.70 \pm 0.23$ & $\phantom{1}6.90 \pm 0.46$ & $\phantom{1}5.55 \pm 0.05$ & $\phantom{1}5.75 \pm 0.05$ & $\phantom{1}\bm{5.21 \pm 0.06}$ & $\phantom{1}5.69 \pm 0.04$ & $\phantom{1}8.36 \pm 0.13$ & $\phantom{1}5.55 \pm 0.01$ \\
$M_{12} (10)$ & $12.99 \pm 0.10$ & $18.81 \pm 0.09$ & $14.00 \pm 0.28$ & $11.90 \pm 0.62$ & $\phantom{1}\bm{9.60 \pm 0.08}$ & $\phantom{1}9.53 \pm 0.08$ & $\phantom{1}9.10 \pm 0.11$ & $\phantom{1}6.76 \pm 0.05$ & $11.03 \pm 0.11$ & $11.61 \pm 0.01$ \\
$M_{13} (20)$ & $25.29 \pm 0.12$ & $35.83 \pm 0.08$ & $34.51 \pm 1.27$ & $20.69 \pm 1.03$ & $16.18 \pm 0.12$ & $15.09 \pm 0.11$ & $15.22 \pm 0.16$ & $\phantom{1}5.85 \pm 0.05$ & $20.88 \pm 0.17$ & $\bm{19.99 \pm 0.00}$ \\
$M_{21} (5)$ & $\phantom{1}8.41 \pm 0.10$ & $\phantom{1}9.65 \pm 0.11$ & $\phantom{1}7.74 \pm 0.22$ & $\phantom{1}7.30 \pm 0.37$ & $\phantom{1}5.58 \pm 0.06$ & $\phantom{1}5.65 \pm 0.05$ & $\phantom{1}5.15 \pm 0.07$ & $\phantom{1}\bm{5.11 \pm 0.04}$ & $\phantom{1}5.72 \pm 0.12$ & $\phantom{1}8.16 \pm 0.01$ \\
$M_{22} (10)$ & $14.78 \pm 0.17$ & $19.35 \pm 0.06$ & $14.40 \pm 0.71$ & $12.01 \pm 0.58$ & $\phantom{1}\bm{9.46 \pm 0.08}$ & $\phantom{1}9.26 \pm 0.07$ & $\phantom{1}8.93 \pm 0.09$ & $\phantom{1}6.25 \pm 0.09$ & $10.70 \pm 0.11$ & $11.78 \pm 0.02$ \\
$M_{23} (20)$ & $25.40 \pm 0.11$ & $36.19 \pm 0.09$ & $35.12 \pm 1.26$ & $20.78 \pm 1.09$ & $16.00 \pm 0.13$ & $14.82 \pm 0.12$ & $15.03 \pm 0.16$ & $\phantom{1}5.67 \pm 0.07$ & $20.36 \pm 0.11$ & $\bm{19.99 \pm 0.00}$ \\
$M_{31} (5)$ & $\phantom{1}5.06 \pm 0.03$ & $\phantom{1}\bm{5.05 \pm 0.02}$ & $\phantom{1}5.97 \pm 0.15$ & $\phantom{1}5.54 \pm 0.26$ & $\phantom{1}4.73 \pm 0.06$ & $\phantom{1}4.90 \pm 0.06$ & $\phantom{1}4.45 \pm 0.07$ & $\phantom{1}4.48 \pm 0.03$ & $\phantom{1}6.68 \pm 0.08$ & $\phantom{1}6.08 \pm 0.01$ \\
\bottomrule
\end{longtable}}
\end{landscape}

\begin{landscape}
{\footnotesize
\setlength{\tabcolsep}{2pt}
\begin{longtable}{@{} l| *{10}{c} @{}}
\caption{Mean dimension estimates ($\pm$ standard deviation) for 18 manifolds (Table B.2) with true dimension $d$, based on 100 replicates of $n=2000$ uniform samples perturbed by Laplace noise. Bold indicates the estimate with minimal MSE.} \label{tab:res08} \\
\toprule
\textbf{Manifold} & \textbf{Local PCA} & \textbf{CA-PCA} & \textbf{DanCo} & \textbf{TwoNN} & \textbf{MLE} & \textbf{TLE} & \textbf{MADA} & \textbf{Wasserstein} & \textbf{QE} & \textbf{TLS} \\
\midrule
\endfirsthead
\multicolumn{11}{c}%
{\tablename\ \thetable\ -- \textit{continued from previous page}} \\
\toprule
\textbf{Manifold} & \textbf{Local PCA} & \textbf{CA-PCA} & \textbf{DanCo} & \textbf{TwoNN} & \textbf{MLE} & \textbf{TLE} & \textbf{MADA} & \textbf{Wasserstein} & \textbf{QE} & \textbf{TLS} \\
\midrule
\endhead
\midrule \multicolumn{9}{r}{\textit{Continued on next page}} \\
\endfoot
\bottomrule
\endlastfoot
$M_{11} (5)$ & $\phantom{1}8.12 \pm 0.04$ & $\phantom{1}9.37 \pm 0.10$ & $\phantom{1}8.03 \pm 0.06$ & $\phantom{1}7.58 \pm 0.21$ & $\phantom{1}6.15 \pm 0.03$ & $\phantom{1}6.30 \pm 0.03$ & $\phantom{1}5.85 \pm 0.03$ & $\phantom{1}5.98 \pm 0.02$ & $\phantom{1}7.33 \pm 0.00$ & $\phantom{1}\bm{5.54 \pm 0.01}$ \\
$M_{12} (10)$ & $14.16 \pm 0.06$ & $19.41 \pm 0.03$ & $14.86 \pm 0.39$ & $12.80 \pm 0.36$ & $10.87 \pm 0.05$ & $10.83 \pm 0.05$ & $\bm{10.40 \pm 0.06}$ & $\phantom{1}7.71 \pm 0.03$ & $13.00 \pm 0.01$ & $10.60 \pm 0.01$ \\
$M_{13} (20)$ & $25.67 \pm 0.06$ & $35.25 \pm 0.04$ & $36.88 \pm 0.89$ & $22.31 \pm 0.56$ & $18.76 \pm 0.07$ & $17.70 \pm 0.07$ & $17.87 \pm 0.09$ & $\phantom{1}6.71 \pm 0.04$ & $\bm{21.00 \pm 0.00}$ & $\bm{21.00 \pm 0.00}$ \\
$M_{21} (5)$ & $\phantom{1}9.35 \pm 0.04$ & $\phantom{1}9.98 \pm 0.00$ & $\phantom{1}9.29 \pm 0.10$ & $\phantom{1}8.11 \pm 0.22$ & $\phantom{1}6.42 \pm 0.03$ & $\phantom{1}6.42 \pm 0.03$ & $\phantom{1}6.04 \pm 0.04$ & $\phantom{1}\bm{5.58 \pm 0.02}$ & $\phantom{1}5.66 \pm 0.01$ & $\phantom{1}8.05 \pm 0.04$ \\
$M_{22} (10)$ & $15.69 \pm 0.07$ & $19.74 \pm 0.02$ & $14.81 \pm 0.26$ & $13.12 \pm 0.34$ & $10.86 \pm 0.05$ & $10.65 \pm 0.05$ & $\bm{10.34 \pm 0.05}$ & $\phantom{1}7.02 \pm 0.05$ & $11.66 \pm 0.01$ & $12.73 \pm 0.01$ \\
$M_{23} (20)$ & $25.84 \pm 0.07$ & $35.60 \pm 0.05$ & $38.13 \pm 0.86$ & $22.41 \pm 0.55$ & $18.68 \pm 0.07$ & $17.47 \pm 0.07$ & $17.77 \pm 0.08$ & $\phantom{1}6.43 \pm 0.05$ & $\bm{20.00 \pm 0.00}$ & $19.99 \pm 0.00$ \\
$M_{31} (5)$ & $\phantom{1}5.78 \pm 0.05$ & $\phantom{1}5.07 \pm 0.02$ & $\phantom{1}7.10 \pm 0.10$ & $\phantom{1}6.24 \pm 0.16$ & $\phantom{1}\bm{5.03 \pm 0.03}$ & $\phantom{1}5.16 \pm 0.02$ & $\phantom{1}4.81 \pm 0.03$ & $\phantom{1}4.91 \pm 0.02$ & $\phantom{1}6.33 \pm 0.01$ & $\phantom{1}6.20 \pm 0.01$ \\
\bottomrule
\end{longtable}}
\end{landscape}

\section*{Appendix E: A Brief Introduction to \textbf{CA-PCA}}
        
\begin{figure}[H]
    \centering
    \begin{subfigure}[b]{0.43\textwidth}
        \centering
        \includegraphics[width=\textwidth]{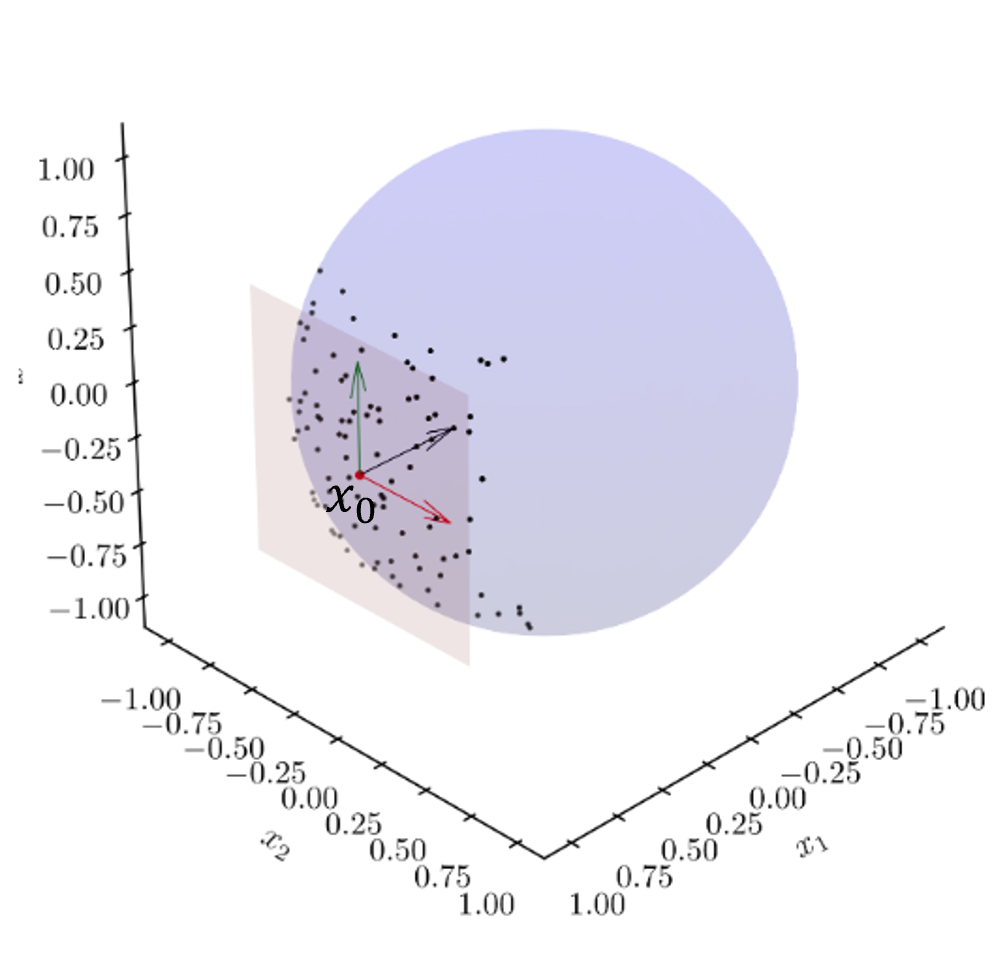}
        \caption{Linear Approximation around $\bm{x}_0$.}
    \end{subfigure}
    \hfill
    \begin{subfigure}[b]{0.45\textwidth}
        \centering
        \includegraphics[width=\textwidth]{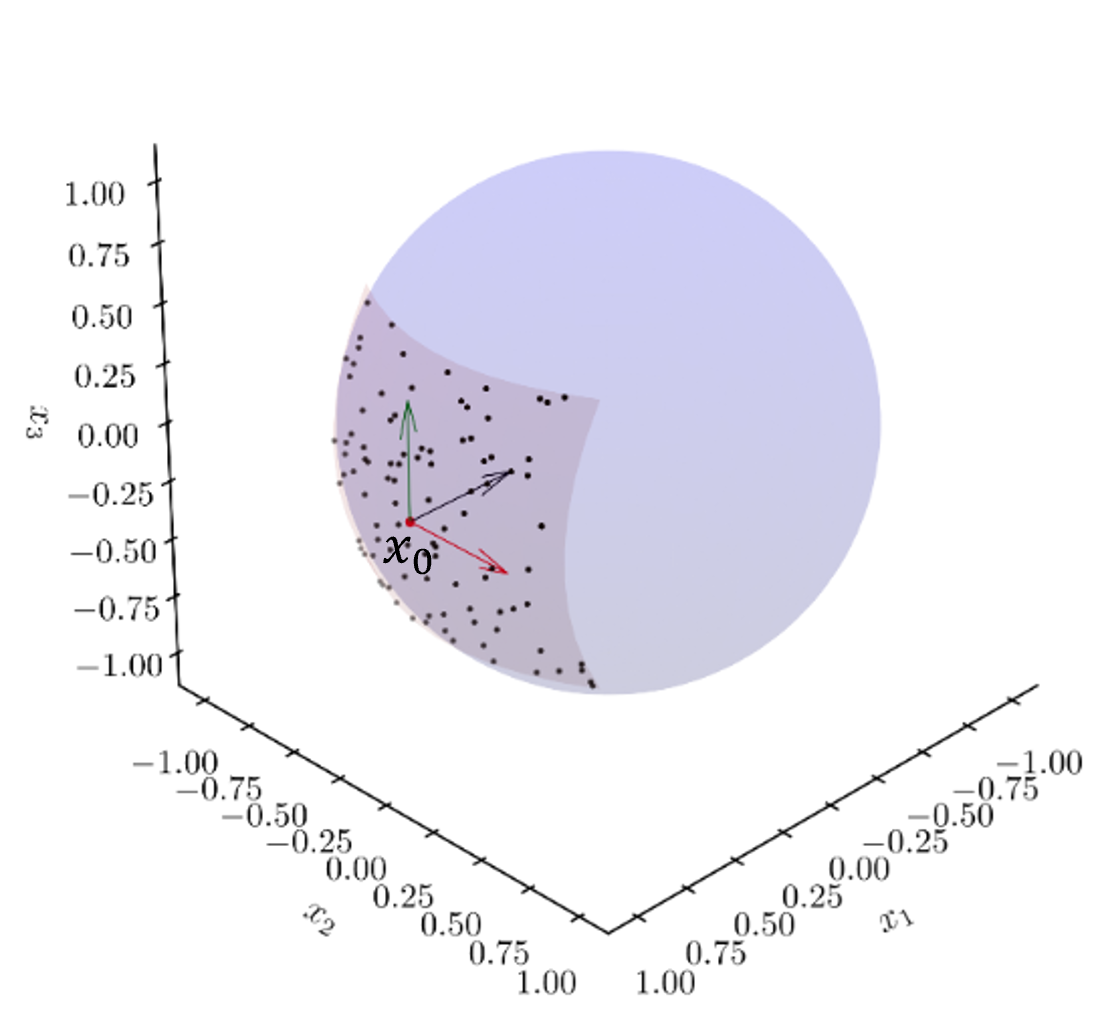}
        \caption{Quadratic Approximation around $\bm{x}_0$.}
    \end{subfigure}
    \caption{Local Graph Representation.}
    \label{fig:mh}
\end{figure}

One formulation of the local PCA estimator is constructed with the eigenvalues of the covariance matrix $\hat{\Sigma}_k$ computed from $\bm{x}_k$ and its neighbors:
\[
\hat{d}_k = \underset{1\leq q\leq p}{\argmin} \sum_{j = 1}^p \lb \frac{1}{R^2}\hat{\lambda}_{j} - \frac{1}{j + 2}1_{j \leq q} \rb^2,
\]
which is based on the fact that under the flatness assumption (within the unit ball), the true covariance matrix $\Sigma_k$ has its first $d$ eigenvalues equal to $1/(d + 2)$ and the remaining $p - d$ eigenvalues equal to $0$ \citep{lim2024tangent}. The main idea of \textbf{CA-PCA} is to adjust the eigenvalues to account for curvature. By the local graph representation of manifolds, instead of assuming that the $\bm{x}_k$ and its neighbors are uniformly distributed in a $d$-dimensional ball in the tangent space, we assume they are uniformly distributed in the \textit{tangent quadratic space} defined through the quadratic approximation of the local graph structure, as illustrated in Figure~\ref{fig:mh}.
It can be shown \citep{gilbert2025pca} that, under the quadratic approximation, the theoretical eigenvalues satisfy
\begin{equation*}
\frac{1}{d + 2} \doteq \lambda_j  + \frac{3d + 4}{d(d + 4)} \sum_{i = d + 1}^p \lambda_i.
\end{equation*}

Using this property, \textsc{CA-PCA} replaces the true eigenvalues with their empirical counterparts and estimates the intrinsic dimension as
\begin{equation*}
\hat{d}_k = \underset{1\leq q\leq p}{\argmin} \left\{\sqrt{\sum_{j=1}^q \left( \frac{1}{q + 2} - \frac{1}{R^2}\lb \hat{\lambda}_{j} + \frac{3q + 4}{q(q + 4)} \sum_{i = q + 1}^p \hat{\lambda}_{ki} \rb \right)^2} + \frac{2}{R^2} \sum_{j = q + 1}^p \hat{\lambda}_{j} \right\},
\end{equation*}
where $R$ can be estimated using $\|\bm{x}_k^K - \bm{x}_k\|$, and the last term serves as a penalty for neighborhoods where the quadratic approximation is of poor quality. A global \textbf{CA-PCA} estimate is then obtained by averaging the local estimates $\hat{d}_k$ across the dataset.

\section*{Appendix F: Estimation Consistency on Real-world Datasets}
\setcounter{table}{0}
\renewcommand{\thetable}{F.\arabic{table}}
\begin{table}[htbp]
\setlength{\tabcolsep}{3pt}
\centering
\caption{Matrix of rank correlation among estimators.}
\label{tab:agreement}
\small
\begin{tabular}{lccccccccc}
\toprule
 & {\footnotesize\textbf{Local PCA}} & {\footnotesize\textbf{CA-PCA}} & {\footnotesize\textbf{TwoNN}} & {\footnotesize\textbf{MLE}} & {\footnotesize\textbf{TLE}} & {\footnotesize\textbf{MADA}} & {\footnotesize\textbf{Wasserstein}} & {\footnotesize\textbf{QE}} & {\footnotesize\textbf{TLS}} \\
\midrule
{\footnotesize\textbf{Local PCA}} & 1.000 & 0.543 & 0.829 & 0.943 & 0.943 & 0.943 & 0.943 & 0.943 & 0.886 \\
{\footnotesize\textbf{CA-PCA}} & 0.543 & 1.000 & 0.829 & 0.657 & 0.657 & 0.657 & 0.657 & 0.600 & 0.486 \\
{\footnotesize\textbf{TwoNN}} & 0.829 & 0.829 & 1.000 & 0.943 & 0.943 & 0.943 & 0.943 & 0.771 & 0.657 \\
{\footnotesize\textbf{MLE}} & 0.943 & 0.657 & 0.943 & 1.000 & 1.000 & 1.000 & 1.000 & 0.886 & 0.829 \\
{\footnotesize\textbf{TLE}} & 0.943 & 0.657 & 0.943 & 1.000 & 1.000 & 1.000 & 1.000 & 0.886 & 0.829 \\
{\footnotesize\textbf{MADA}} & 0.943 & 0.657 & 0.943 & 1.000 & 1.000 & 1.000 & 1.000 & 0.886 & 0.829 \\
{\footnotesize\textbf{Wasserstein}} & 0.943 & 0.657 & 0.943 & 1.000 & 1.000 & 1.000 & 1.000 & 0.886 & 0.829 \\
{\footnotesize\textbf{QE}} & 0.943 & 0.600 & 0.771 & 0.886 & 0.886 & 0.886 & 0.886 & 1.000 & 0.943 \\
{\footnotesize\textbf{TLS}} & 0.886 & 0.486 & 0.657 & 0.829 & 0.829 & 0.829 & 0.829 & 0.943 & 1.000 \\
\bottomrule
\end{tabular}
\end{table}

\section*{Appendix G: Runtime Ratios}

All experiments are conducted with datasets sampled from the unit sphere.

\setcounter{figure}{0}
\renewcommand{\thefigure}{G.\arabic{figure}}

\begin{figure}[H]
\centering
\includegraphics[width=\textwidth]{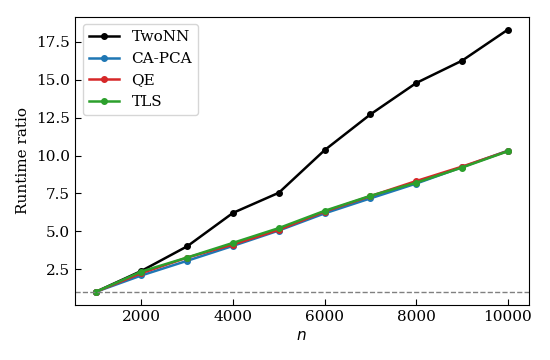}
\caption{Runtime ratios of different estimators as a function of the sample size $n$.}
\label{fig:ratio_n}
\end{figure}

\begin{figure}[H]
\centering
\includegraphics[width=\textwidth]{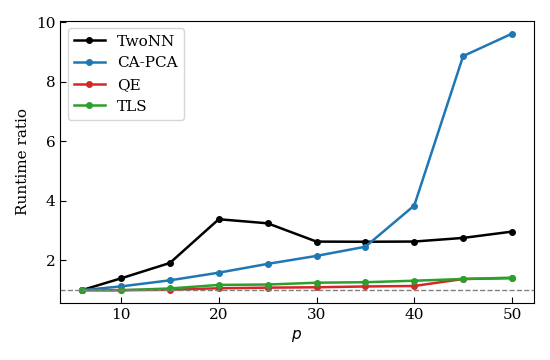}
\caption{Runtime ratios of different estimators as a function of the ambient dimension $p$.}
\label{fig:ratio_p}
\end{figure}

\begin{figure}[H]
\centering
\includegraphics[width=\textwidth]{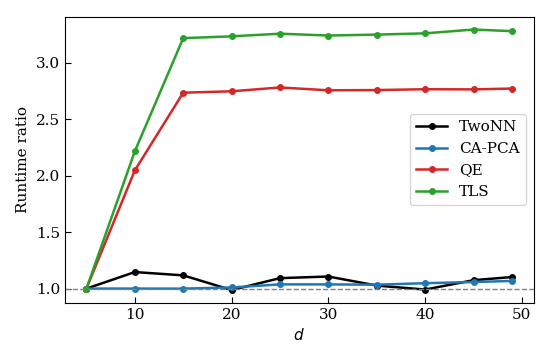}
\caption{Runtime ratios of different estimators as a function of the intrinsic dimension $d$.}
\label{fig:ratio_d}
\end{figure}

\section*{Appendix H: Proofs of Theoretical Properties}

\setcounter{theorem}{0}
\setcounter{proposition}{0}

Here, we provide detailed descriptions and proofs of the theoretical properties introduced in Section 3.3 of the manuscript. As discussed there, rather than seeking a complete consistency theory, we focus on a local neighborhood and establish these properties under an approximate local model. Since both \textbf{QE} and \textbf{TLS} apply their procedures across many local neighborhoods and then aggregate the resulting estimates, their overall performance is closely tied to the behavior of the estimators at the local level. Thus, the local theoretical properties provide an important foundation for understanding their empirical performance over the full dataset.

Consider a point $\bm{x}_0 \in M$ and $K$ neighboring points
$\bm{x}_0^{1},\ldots,\bm{x}_0^{K}$ from the dataset around it. Let
$\bm{x}_1',\ldots,\bm{x}_K'$ denote their coordinates with respect to \textit{any} orthonormal basis centered at $\bm{x}_0$, whose first $d$ directions span the tangent space $T_{\bm{x}_0}M$. Under the manifold hypothesis and the local graph representation property of the underlying manifold, when the local scale is sufficiently small, say when all neighboring points lie within $B_{\bm{x}_0}(h)$ for some $h>0$, the local distribution of the data can be approximately modeled as
\begin{equation}
    \bm{x}'
    =
    \begin{pmatrix}
        h\bm{z} \\
        \bm{q}_d(h\bm{z})
    \end{pmatrix}
    +
    \bm{\varepsilon},
    \label{eq:local_model}
\end{equation}
where $\bm{z}\in\mathbb{R}^d$ is uniformly distributed within the unit ball, $\bm{\varepsilon}\sim N(0,\sigma^2I_p)$ is independent
noise, and
\begin{equation}
    q_{d\ell}(h\bm{z})
    =
    h^2\bm{z}^{\top}Q_{d\ell}\bm{z},
    \qquad
    \ell=1,\ldots,p-d,
\end{equation}
with symmetric matrices
$Q_{d\ell}\in\mathbb{R}^{d\times d}$. 

To justify the estimators, we establish the following results with respect to the local model above.

Proposition~\ref{prop:population-pca} below shows that, under the local model and a regular condition, population PCA recovers the tangent space. This provides theoretical justification for using PCA to estimate tangent spaces in both \textbf{QE} and \textbf{TLS}.

\begin{proposition}
Let
\[
    \Gamma(h)
    =
    \operatorname{Var}\!\left[
        \bm{q}_d(h\bm{z})
    \right].
\]
Then the covariance matrix of $\bm{x}'$ has $d$ tangent eigenvalues
\[
    \lambda_j
    =
    \sigma^2+\frac{h^2}{d+2},
    \qquad
    j=1,\ldots,d,
\]
and $p-d$ normal eigenvalues
\[
    \lambda_j
    =
    \sigma^2+h^4\mu_{j-d},
    \qquad
    j=d+1,\ldots,p,
\]
where
\[
    \mu_1\geq\cdots\geq\mu_{p-d}\geq0
\]
are the eigenvalues of
\[
    \Gamma(1)
    =
    \operatorname{Var}\!\left[
        \bm{q}_d(\bm{z})
    \right].
\]
In particular, if
\begin{equation}
    \frac{h^2}{d+2}>h^4\mu_1,
    \label{eq:eigengap}
\end{equation}
then the first $d$ population principal components span the tangent space
$T_{\bm{x}_0}M$.
\end{proposition}

\begin{proof}
Since $\bm{z}$ is uniformly distributed on the unit ball in
a $d$-dimensional space, it follows from standard results \citep{fang1990} that
\[
    \mathbb{E}[\bm{z}]=\bm{0},
    \qquad
    \operatorname{Cov}(\bm{z})
    =
    \frac{1}{d+2}I_d.
\]
Moreover, since each component of $\bm{q}_d(h\bm{z})$ is quadratic in
$\bm{z}$,
\[
    \mathbb{E}\!\left[
        \bm{z}\bm{q}_d(h\bm{z})^\top
    \right]
    =
    \bm{0},
\]
because the integrand is an odd function of $\bm{z}$ over a centrally
symmetric domain. Therefore, the tangent and normal components in
\eqref{eq:local_model} are uncorrelated.

Writing
\[
    \bm{x}'
    =
    \begin{pmatrix}
        h\bm{z}\\
        \bm{q}_d(h\bm{z})
    \end{pmatrix}
    +
    \bm{\varepsilon},
\]
and using the independence of $\bm{\varepsilon}$, we obtain
\[
    \operatorname{Cov}(\bm{x}')
    =
    \begin{pmatrix}
        h^2\operatorname{Cov}(\bm z) & \bm 0\\
        \bm 0 & \Gamma(h)
    \end{pmatrix}
    +
    \sigma^2 I_p.
\]
Consequently,
\[
    \operatorname{Cov}(\bm{x}')
    =
    \begin{pmatrix}
        \left(\sigma^2+\dfrac{h^2}{d+2}\right)I_d
        & \bm 0\\[6pt]
        \bm 0
        & \sigma^2 I_{p-d}+\Gamma(h)
    \end{pmatrix}.
\]

It remains to characterize the normal block. Since
\[
    \bm{q}_d(h\bm z)
    =
    h^2\bm{q}_d(\bm z),
\]
we have
\[
    \Gamma(h)
    =
    \operatorname{Var}\!\left[
        h^2\bm{q}_d(\bm z)
    \right]
    =
    h^4\Gamma(1).
\]
Thus, if
\[
    \mu_1\geq\cdots\geq\mu_{p-d}\geq0
\]
are the eigenvalues of $\Gamma(1)$, the $p - d$ eigenvalues corresponding to normal directions
are
\[
    \sigma^2+h^4\mu_1,\ldots,
    \sigma^2+h^4\mu_{p-d}.
\]
Together with the $d$ eigenvalues corresponding to tangent directions
\[
    \sigma^2+\frac{h^2}{d+2},
\]
this gives the claimed eigenvalue decomposition.

Finally, under the eigengap condition
\[
    \frac{h^2}{d+2}>h^4\mu_1,
\]
we have
\[
    \sigma^2+\frac{h^2}{d+2}
    >
    \sigma^2+h^4\mu_1
    \geq
    \sigma^2+h^4\mu_j,
    \qquad
    j=1,\ldots,p-d.
\]
Hence, the $d$ largest eigenvalues of $\operatorname{Cov}(\bm{x}')$
are precisely the tangent eigenvalues. Their associated eigenspace is
therefore the first $d$ population principal component subspace, which
coincides with $T_{\bm{x}_0}M$.
\end{proof}

Proposition~\ref{prop:first-normal-component} below further shows that, if $M$ has nonzero curvature at $\bm{x}_0$, then the $(d+1)$-st coordinate from population PCA contains a nonconstant quadratic function of the first $d$ PCA coordinates. This provides a theoretical foundation for why quadratically regressing the $(d+1)$-st coordinate on the first $d$ coordinates is appropriate.

\begin{proposition}
Suppose that the matrices $Q_{d\ell}$ are not all zero. Then
\[
    \mu_1 > 0.
\]
Under the eigengap condition \eqref{eq:eigengap}, the $(d+1)$-st
population principal component is a nonzero normal direction whose coordinate contains a nonconstant quadratic function of
the first $d$ tangent coordinates.
\end{proposition}

\begin{proof}
Since the matrices $Q_{d\ell}$ are not all zero,
$\bm{q}_d(\bm{z})$ is not identically zero. Hence,
\[
    \Gamma(1)
    =
    \operatorname{Var}\!\left[
        \bm{q}_d(\bm{z})
    \right]
    \neq 0.
\]
As $\Gamma(1)$ is positive semidefinite, its largest eigenvalue satisfies
\[
    \mu_1 > 0.
\]

Let $\bm{v}_1$ be a unit eigenvector of $\Gamma(1)$ associated with
$\mu_1$. Since
\[
    \Gamma(h) = h^4\Gamma(1),
\]
we have
\[
    \left(
        \sigma^2 I_{p-d} + \Gamma(h)
    \right)\bm{v}_1
    =
    \left(
        \sigma^2 + h^4\mu_1
    \right)\bm{v}_1.
\]
Therefore,
\[
    \begin{pmatrix}
        \bm{0}\\
        \bm{v}_1
    \end{pmatrix}
\]
is an eigenvector of $\operatorname{Cov}(\bm{x}')$ with eigenvalue
$\sigma^2 + h^4\mu_1$. Under the eigengap condition
\eqref{eq:eigengap}, this eigenvalue is smaller than the $d$ tangent
eigenvalues. Hence, the corresponding eigenvector is the $(d+1)$-st
population principal component.

Its corresponding noiseless coordinate is
\[
    \bm{v}_1^\top \bm{q}_d(h\bm{z})
    =
    h^2 \bm{v}_1^\top \bm{q}_d(\bm{z})
    =
    h^2\bm{z}^\top Q_1\bm{z},
\]
where
\begin{equation}\label{Q1}
    Q_1
    =
    \sum_{\ell=1}^{p-d}v_{1\ell}Q_{d\ell}.
\end{equation}
Since $\bm{v}_1^\top\bm{q}_d(\bm{z})$ is a nonconstant quadratic
function of $\bm{z}$, we have $Q_1\neq0$.
\end{proof}

Propositions \ref{prop:population-pca} and
\ref{prop:first-normal-component} hold for any orthonormal basis that
gives rise to $\bm{x}'$. Since any two such choices of basis differ only
by orthogonal transformations within the tangent and normal spaces, we
may, without loss of generality, assume that the initial orthonormal
basis coincides with the basis induced by population PCA, which can greatly simplify the proofs of our main result.

Next, we establish our main result after presenting some preliminaries. 

Recall first that a sequence of random variables $\{X_K\}$ is bounded in probability, denoted by $X_K = O_{\mathbb{P}}(1)$, if, for every $\epsilon > 0$, there exists $M > 0$ such that
\[
    \limsup_{K \to \infty}
    \mathbb{P}\!\left(
        |X_K| > M
    \right)
    < \epsilon,
\]
i.e., the probability of $X_K$ taking arbitrarily large values becomes negligible as $K$ increases.

Also, for candidate dimension $j \geq 1$, let
\begin{equation}
    \mathcal{S}_j
    =
    \operatorname{span}
    \left\{
        1,\,
        x'_1,\ldots,x'_j,\,
        x'_a x'_b:
        1\leq a\leq b\leq j
    \right\}
\end{equation}
be the feature space, and denote by $m_j$ the number of nonconstant regressors,
\begin{equation}
    m_j
    =
    j+\binom{j+1}{2}.
\end{equation}
The \textbf{QE} procedure regresses $x'_{j+1}$ on the generating elements of $\mathcal{S}_j$ and uses the corresponding overall $F$-statistic. Define
\begin{equation}
    \bm{\phi}_{j}(\bm{x}_{i,1:j}')
    =
    \left(
        x'_{i1},\ldots,x'_{ij},
        \left\{
            x'_{ia}x'_{ib}:1\leq a\leq b\leq j
        \right\}
    \right)^\top
    \in\mathbb{R}^{m_j},
\end{equation}
as the vector of nonconstant regressors generated from $\bm{x}'_i$. We further define their sample mean and the sample mean of the response by
\begin{equation}
    \overline{\bm{\phi}}_{j}
    =
    \frac{1}{K}
    \sum_{i=1}^K
    \bm{\phi}_{j}(\bm{x}_{i,1:j}'),
    \qquad
    \overline{x}_{j+1}'
    =
    \frac{1}{K}
    \sum_{i=1}^K x'_{i,j+1}.
\end{equation}
The OLS coefficient estimator can then be written in terms of the centered regressors and response as
\begin{equation}
    \widehat{\bm\beta}_{j,K}
    =
    \left[
        \sum_{i=1}^K
        \left(
            \bm{\phi}_{j}(\bm{x}_{i,1:j}')
            -\overline{\bm{\phi}}_{j}
        \right)
        \left(
            \bm{\phi}_{j}(\bm{x}_{i,1:j}')
            -\overline{\bm{\phi}}_{j}
        \right)^\top
    \right]^{-1}
    \sum_{i=1}^K
        \left(
            \bm{\phi}_{j}(\bm{x}_{i,1:j}')
            -\overline{\bm{\phi}}_{j}
        \right)
        \left(
            x'_{i,j+1}-\overline{x}_{j+1}'
        \right).
\end{equation}

Furthermore, the overall $F$-statistic is computed as
\begin{equation}\label{F_jk}
    F_{j,K}
    =
    \frac{\mathrm{SSR}_{j,K}/m_j}
         {\mathrm{SSE}_{j,K}/(K-m_j-1)} =
    \frac{\widehat R_{j,K}^2/m_j}
         {(1-\widehat R_{j,K}^2)/(K-m_j-1)},
\end{equation}
where
\begin{equation}\label{ssr}
    \mathrm{SSR}_{j,K}
    =
    \widehat{\bm\beta}_{j,K}^{\top}
    \left[
        \sum_{i=1}^K
        \left(
            \bm{\phi}_{j}(\bm{x}_{i,1:j}')
            -\overline{\bm{\phi}}_{j}
        \right)
        \left(
            \bm{\phi}_{j}(\bm{x}_{i,1:j}')
            -\overline{\bm{\phi}}_{j}
        \right)^\top
    \right]
    \widehat{\bm\beta}_{j,K},
\end{equation}
\begin{equation}
    \mathrm{SSE}_{j,K}
    =
    \sum_{i=1}^K
    \left[
        x'_{i,j+1}
        -
        \overline{x}_{j+1}'
        -
        \widehat{\bm\beta}_{j,K}^{\top}
        \left(
            \bm{\phi}_{j}(\bm{x}_{i,1:j}')
            -
            \overline{\bm{\phi}}_{j}
        \right)
    \right]^2,
\end{equation}
are regression and residual sums of squares, respectively, and $\widehat R_{j,K}^2$ is the sample coefficient of determination. Finally, define
\begin{equation}
    \rho_h^2
    =
    \frac{
        \operatorname{Var}\!\left(
            \operatorname{proj}_{\mathcal{S}_d} x'_{d+1}
        \right)
    }{
        \operatorname{Var}(x'_{d+1})
    },
    \label{eq:rho-definition}
\end{equation}
as the population coefficient of determination of the quadratic regression under the local model, which is the quantity $\widehat R_{d,K}^2$ seeks to estimate.

Consider the asymptotic regime in which the local approximate model remains valid within $B_{\bm{x}_0}(h)$ while the number of neighboring points $K$ increases with the sample size, our main result, Theorem~\ref{thm:qe-separation} below, shows that the $F$-statistic used by \textbf{QE} to test candidate dimension $j$, denoted by $F_{j,K}$, satisfies
\[
    F_{j,K} = O_{\mathbb{P}}(1),
    \qquad j < d,
\]
while
\[
    \frac{F_{d,K}}{K} = O_{\mathbb{P}}(1).
\]
Thus, the $F$-statistics for dimensions below $d$ remain bounded in probability, whereas the statistic at the true dimension is of order $K$. As $K \to \infty$, this provides asymptotic separation between the true intrinsic dimension and all lower candidate dimensions, consistent with the separation observed empirically in Figure~4 of the manuscript.

\begin{theorem}

Let $F_{j,K}$ denote the $F$-test statistic of the \textbf{QE} estimator
for candidate dimension $j$. Then, for $j=1,\ldots,d-1$,
\begin{equation}
    F_{j,K}
    =
    O_{\mathbb{P}}(1),
    \label{eq:F-sub-d}
\end{equation}
whereas
\begin{equation}
    \frac{F_{d,K}}{K}
    \xrightarrow{\mathbb{P}}
    \frac{\rho_h^2}
         {m_d(1-\rho_h^2)},
    \qquad
    K\to\infty,
    \label{eq:F-at-d}
\end{equation}
where $\rho_h^2 \in (0, 1)$. Thus, the \textbf{QE} procedure asymptotically distinguishes
the true intrinsic dimension $d$ from every lower candidate dimension:
for $j<d$, $F_{j,K}$ remains bounded in probability, whereas
$F_{d,K}$ grows linearly with $K$.
\end{theorem}

\begin{proof}
By the comment above, we may work in the orthonormal basis induced by
population PCA. In this coordinate system, the first $d$ coordinates are
the tangent coordinates, while the $(d+1)$-st coordinate is the first
normal principal component. For each candidate dimension $j \geq 1$, the \textbf{QE} procedure regresses $x'_{j+1}$ on the elements of $\mathcal{S}_j$ and computes $F_{j,K}$. We study the asymptotic properties of $F_{j,K}$ for $j<d$ and $j=d$.

We first consider $j<d$. Under the local model,
\[
    x'_a = hz_a+\varepsilon_a,
    \qquad a=1,\ldots,d,
\]
where $\bm z$ is uniformly distributed on the unit ball and
$\bm\varepsilon\sim N(0,\sigma^2I_p)$ is independent of $\bm z$.
By the reflection symmetry $z_{j+1}\mapsto-z_{j+1}$, conditional on
$z_1,\ldots,z_j$, we have
\[
    \mathbb{E}[z_{j+1}\mid z_1,\ldots,z_j]=0.
\]
The same symmetry, together with the independence and centering of the
noise, yields
\[
    \mathbb{E}[x'_{j+1}\mid\bm{x}'_{1:j}]=0.
\]
Consequently,
\[
    \mathbb{E}\!\left[
        x'_{j+1}g(\bm{x}'_{1:j})
    \right]=0
\]
for every square-integrable function $g$. Hence, $x'_{j+1}$ is orthogonal in $L_2$ to every element in
$\mathcal{S}_j$. Therefore, all population regression coefficients are zero, and the corresponding population
coefficient of determination satisfies
\[
    R_j^2=0,
    \qquad j=1,\ldots,d-1.
\]

Let $\widehat{\bm\beta}_{j,K}$ denote the vector of fitted
coefficients corresponding to the $m_j$ nonconstant regressors in
$\mathcal{S}_j$. Since the population regression
coefficients are all zero and the relevant moments are finite under our model, the
multivariate central limit theorem implies
\[
    \frac{1}{\sqrt{K}}
    \sum_{i=1}^K
        \left(
            \bm{\phi}_{j}(\bm{x}_{i,1:j}')
            -\overline{\bm{\phi}}_{j}
        \right)
        \left(
            x_{j+1}^{\prime i}-\overline{x}_{j+1}'
        \right)
    =
    O_{\mathbb P}(1).
\]
Thus, the sample cross-product in the OLS estimator is
$O_{\mathbb P}(\sqrt{K})$.

On the other hand, the law of large numbers gives
\[
    \frac{1}{K}
    \sum_{i=1}^K
        \left(
            \bm{\phi}_{j}(\bm{x}_{i,1:j}')
            -\overline{\bm{\phi}}_{j}
        \right)
        \left(
            \bm{\phi}_{j}(\bm{x}_{i,1:j}')
            -\overline{\bm{\phi}}_{j}
        \right)^{\top}
    \xrightarrow{\mathbb P}
    \Sigma_j,
\]
where $\Sigma_j$ is the population covariance matrix of the
nonconstant regressors and is positive definite. Hence, the sample
design matrix is of order $K$ and its inverse is of order $K^{-1}$, i.e.,
\[
    \left[
        \sum_{i=1}^K
        \left(
            \bm{\phi}_{j}(\bm{x}_{i,1:j}')
            -\overline{\bm{\phi}}_{j}
        \right)
        \left(
            \bm{\phi}_{j}(\bm{x}_{i,1:j}')
            -\overline{\bm{\phi}}_{j}
        \right)^{\top}
    \right]^{-1}
    =
    O_{\mathbb P}(K^{-1}).
\]
Combining this with the $O_{\mathbb P}(\sqrt{K})$ bound for the
cross-product yields
\[
    \widehat{\bm\beta}_{j,K}
    =
    O_{\mathbb P}(K^{-1/2}).
\]

Hence, according to (\ref{ssr}), the regression sum of squares satisfies
$$
    \mathrm{SSR}_{j,K}
    =
    O_{\mathbb P}(K^{-1/2})
    O_{\mathbb P}(K)
    O_{\mathbb P}(K^{-1/2})
    =
    O_{\mathbb P}(1).
$$
Moreover, since $R_j^2=0$, the population regression residual is
$x'_{j+1}$ itself, and hence
$$
    \frac{\mathrm{SSE}_{j,K}}{K}
    \xrightarrow{\mathbb P}
    \operatorname{Var}(x'_{j+1})>0.
$$
It follows that
$$
    F_{j,K}
    =
    \frac{\mathrm{SSR}_{j,K}/m_j}
         {\mathrm{SSE}_{j,K}/(K-m_j-1)}
    =
    O_{\mathbb P}(1),
    \qquad
    j=1,\ldots,d-1.
$$
This proves \eqref{eq:F-sub-d}.

We next consider $j=d$ case and first show that the population coefficient of determination $\rho_h^2$ is strictly between $0$ and $1$. 

To show that $\rho_h^2>0$, it suffices to find a quadratic function of
$\bm{x}_{1:d}'$ that has nonzero covariance with $x'_{d+1}$. Consider
\[
    {\bm{x}_{1:d}'}^\top Q_1\bm{x}_{1:d}'\in\mathcal{S}_d,
\]
where $Q_1$ is defined as in (\ref{Q1}).
Since
\[
    {\bm{x}_{1:d}'}^\top Q_1\bm{x}_{1:d}'
    =
    h^2\bm z^\top Q_1\bm z
    +2h\bm z^\top Q_1\bm\varepsilon_{1:d}
    +\bm\varepsilon_{1:d}^\top Q_1\bm\varepsilon_{1:d},
\]
the independence and centering of the noise implies
\[
    \operatorname{Cov}
    \left(
        x'_{d+1},
        {\bm{x}_{1:d}'}^\top Q_1\bm{x}_{1:d}'
    \right)
    =
    \frac{2h^4}{(d+2)(d+4)}
    \left[
        \operatorname{tr}(Q_1^2)
        -
        \frac{\operatorname{tr}(Q_1)^2}{d+2}
    \right].
\]
Since $Q_1\neq0$,
\[
    \operatorname{tr}(Q_1)^2
    \leq d\,\operatorname{tr}(Q_1^2)
    <(d+2)\operatorname{tr}(Q_1^2),
\]
so the covariance is strictly positive. Therefore, the population
projection of $x'_{d+1}$ onto $\mathcal{S}_d$ is nonconstant, and hence
$\rho_h^2>0$.

Moreover, since $\varepsilon_{d+1}$ is independent of
$\bm{x}_{1:d}'$ and $\sigma^2>0$, it contributes strictly positive
unexplained variance, and we have the other direction.

Finally, let $\widehat R_{d,K}^2$ be the sample coefficient of
determination of the quadratic regression of $x'_{d+1}$ on
$\mathcal S_d$. Since the number of regressors is fixed and all
variables have finite moments, the sample version should
converge to its population counterpart. Therefore,
$$
    \widehat R_{d,K}^2
    \xrightarrow{\mathbb P}
    \rho_h^2.
$$
Since the overall $F$-statistic for the $m_d$ nonconstant regressors can be expressed as
$$
    F_{d,K}
    =
    \frac{\widehat R_{d,K}^2/m_d}
         {(1-\widehat R_{d,K}^2)/(K-m_d-1)}.
$$
Consequently,
$$
    \frac{F_{d,K}}{K}
    =
    \frac{K-m_d-1}{K}
    \frac{\widehat R_{d,K}^2}
         {m_d(1-\widehat R_{d,K}^2)}
    \xrightarrow{\mathbb P}
    \frac{\rho_h^2}
         {m_d(1-\rho_h^2)},
$$
i.e., the claimed result \eqref{eq:F-at-d} follows. Thus, for every
$j<d$, the \textbf{QE} statistic remains bounded in probability, whereas at the
true intrinsic dimension it grows linearly with $K$, establishing the
asymptotic separation.
\end{proof}


Finally, note that in practice, both $\bm{x}_0$ and the population PCA coordinates are not available. We therefore apply the procedure locally around each observed point $\bm{x}_k$, performing PCA on $\bm{x}_k$ and its $K$ nearest neighbors as an approximation to the population PCA. The unknown center $\bm{x}_0$ is approximated by the corresponding sample mean.

\bibliographystyle{elsarticle-num} 
\bibliography{reference}






\end{document}